%% file: main.tex
\documentclass[10pt,twocolumn,letterpaper]{article}

\usepackage[pagenumbers]{wacv} 
\usepackage{graphicx}
\usepackage{amsmath}
\usepackage{amssymb}
\usepackage{booktabs}
\usepackage{adjustbox}

\usepackage{times}
\usepackage{epsfig}
\usepackage[table]{xcolor}
\usepackage{tikz}
\usepackage{tabularx}
\usepackage{cuted}
\usepackage{textcomp}
\usepackage[ruled,vlined]{algorithm2e}
\usepackage{multirow}
\usepackage{anyfontsize}
\usepackage{amsmath}
\usepackage{amsfonts}
\usepackage{bm}
\usepackage{comment}

\input{math-commands}

\usepackage[pagebackref,breaklinks,colorlinks]{hyperref}

\usepackage[capitalize]{cleveref}
\crefname{section}{Sec.}{Secs.}
\Crefname{section}{Section}{Sections}
\Crefname{table}{Table}{Tables}
\crefname{table}{Tab.}{Tabs.}

\def\wacvPaperID{59} 
\def\confName{WACV}
\def\confYear{2024}

\begin{document}

\title{Object Re-Identification from Point Clouds}

\author{Benjamin Th\'erien \qquad  Chengjie Huang \qquad Adrian Chow \qquad Krzysztof Czarnecki\\
University of Waterloo\\
{\tt\small \{btherien,c.huang,adrian.hei.tung.chow,k2czarne\}@uwaterloo.ca}}

\maketitle

\begin{abstract}

Object re-identification (ReID) from images plays a critical role in application domains of image retrieval (surveillance, retail analytics, etc.) and multi-object tracking (autonomous driving, robotics, etc.). However, systems that additionally or exclusively perceive the world from depth sensors are becoming more commonplace without any corresponding methods for object ReID. In this work, we fill the gap by providing the first large-scale study of object ReID from point clouds and establishing its performance relative to image ReID. To enable such a study, we create two large-scale ReID datasets with paired image and LiDAR observations and propose a lightweight matching head that can be concatenated to any set or sequence processing backbone (e.g., PointNet or ViT), creating a family of comparable object ReID networks for both modalities. Run in Siamese style, our proposed point cloud ReID networks can make thousands of pairwise comparisons in real-time ($10$\,Hz). Our findings demonstrate that their performance increases with higher sensor resolution and approaches that of image ReID when observations are sufficiently dense. Our strongest network trained at the largest scale achieves ReID accuracy exceeding $90\%$ for rigid objects and $85\%$ for deformable objects (without any explicit skeleton normalization). To our knowledge, we are the first to study object re-identification from real point cloud observations.

\end{abstract}

\section{Introduction}


Re-identification from images is a core component in many application domains such as surveillance\cite{zheng2019indentity}, retail analytics \cite{wei2020retail}, autonomous driving\cite{kim2021eagermot,willes2022intertrack,weng2020gnn3dmot,wangcamomot2021}, robotics \cite{Zheng2018picknplace}, and many more. Given the increasing deployment of high-resolution LiDAR sensors \cite{sun2020waymo,ceasar2020nuscenes,chang2019argoverse}, especially as part of robot perception systems, the development of similar techniques for ReID from point clouds has the potential to enhance these systems with a host of new capabilities. Among them, appearance-based re-identification for multi-object tracking is, perhaps, the most impactful. For instance, in robotics, whether for navigation in complex environments or for tasks like pick-and-place, the ability to accurately identify and track multiple objects in 3D space is crucial. Moreover, multi-object tracking is essential for the safe operation of autonomous vehicles.

\begin{figure*}[ht]
    \centering
    \subfloat{{\includegraphics[width=0.49\linewidth]{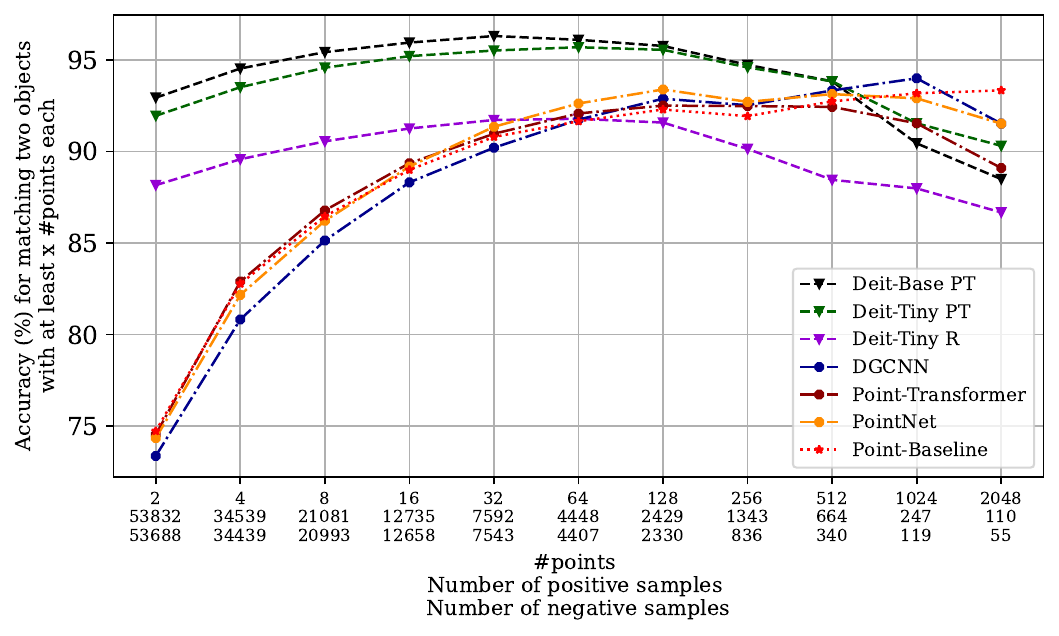} }}
    \subfloat{{\includegraphics[width=0.49\linewidth]{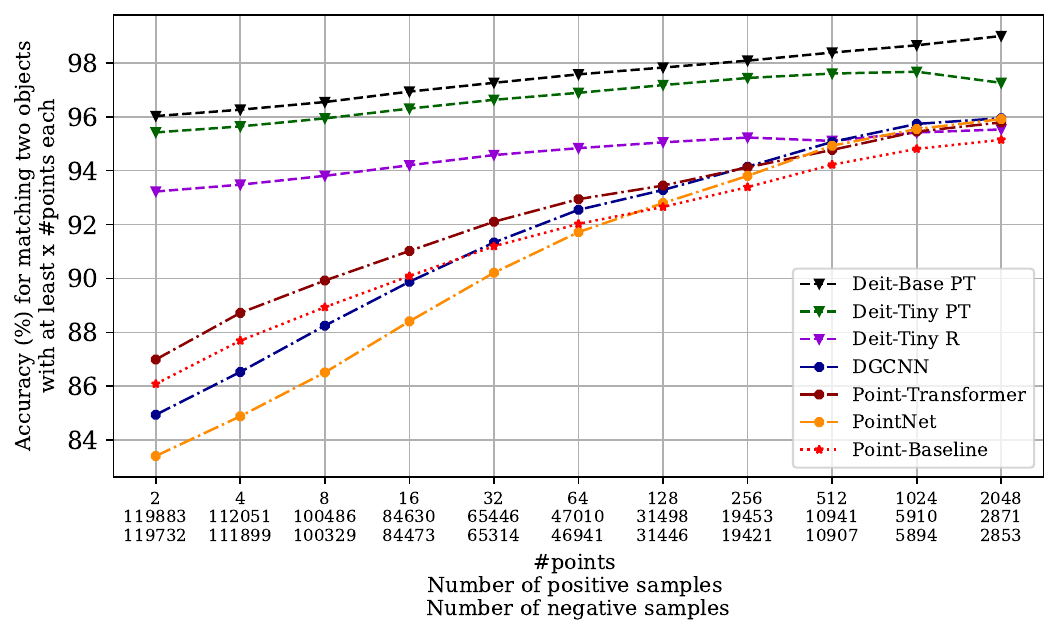} }}
    \vspace{-8pt}
    \caption{\textbf{The performance of point cloud ReID approaches image ReID with sufficient points.} We plot the performance of image and point cloud ReID networks as a function of point density. Left shows models trained on nuScenes and evaluated on \emph{nuScenes Eval} set, while right shows models trained on the Waymo Open Dataset (WOD) and evaluated on \emph{Waymo Eval}. }
    \vspace{-13pt}
    \label{fig:pointreidapproaches}
\end{figure*}

While strong ReID performance can be obtained from image data alone, even autonomous agents equipped with arrays of RGB sensors stand to benefit from the added redundancy, diversity, and complementarity offered by processing depth-sensor information for ReID. Despite this clear added benefit, however, the existing literature on re-identification from point cloud data is almost non-existent. To the best of our knowledge, \cite{zheng20223dreid} is the only work directly studying the problem, but they do so on a synthetic person re-identification dataset. Together with the clear motivation for leveraging ReID from point clouds for many applications, the lack of established knowledge about this problem 
motivates our central research question: \emph{how effective is LiDAR-based ReID compared to camera-based ReID?} 

While LiDAR sensors are devoid of the lighting challenges that affect cameras, they have unique challenges of their own. The primary difficulty is the sparsity of LiDAR scans and the lack of color and texture information compared to images. A network trained to re-identify objects from point clouds must rely solely on shape information. However, certain deformable objects can pose particular difficulty as their pose can change over time. An open question is whether reliable re-identification of pedestrians is even possible from raw LiDAR input without using explicit normalization schemes. Although such questions are simple, the lack of readily available high-quality datasets for re-identification from point clouds has been a serious impediment to research thus far. We address it in what follows by providing a simple recipe for creating point clouds re-identification datasets from large-scale autonomous driving datasets.

To the best of our knowledge, we are the first work to investigate object re-identification from point cloud observations.  
Our contributions can be summarized as follows:
\begin{itemize}
    \item We propose RTMM, a symmetric matching head for ReID from point clouds that runs in real-time and shows improved convergence and generalization compared to a strong baseline. 
    \item We provide a recipe for creating re-identification datasets from large-scale autonomous driving datasets and propose a performant training-time sampling algorithm.
    \item We are the first to establish point cloud ReID performance relative to image ReID on large-scale datasets. Our results demonstrate that point cloud ReID can approach the performance of image ReID when LiDAR observations are sufficiently dense.
    \item We fit a power-law to estimate the benefits of training beyond our compute budget, suggesting that an improvement of $~2\%$ above our strongest ReID model ($89.3$\% ReID accuracy) is attainable with an order of magnitude more compute.

\end{itemize}

Our results outline a promising future for point-based object ReID, especially as depth-sensor resolution continues to increase. Upon publication, we will release the code used to collect our re-identification datasets and train our networks. 

\section{Related Work}

This section briefly outlines areas relevant to our study of object ReID from point clouds.  

\subsection{Point-Processing Networks}
The ability to effectively represent irregular sets of points is essential for 3D geometry processing. Respecting the symmetries of permutation invariance (PointNet)~\cite{qipointnet2017} and the metric space structure of raw point clouds (PointNet++)~\cite{qi2017pnpp} were shown early on to be important priors. Subsequent works propose edge convolutions to process point clouds in CNN-style \cite{wang2019dgcnn},  a performant and efficient residual-MLP framework \cite{ma2022residualmlp}, exploiting the benefits of depth \cite{le2020lean}, using the MLP-Mixer~\cite{choe2022pointmixer}, using a transformer-based architecture \cite{zhao2021pointtransformer}, among others. In the following study, we select three efficient models to use for our experiments: PointNet~\cite{qipointnet2017}, DGCNN~\cite{wang2019dgcnn}, and Point-Transformer~\cite{hui2022siamesepointtrans,zhao2021pointtransformer}.
 
\begin{figure*}[ht]
    \centering
    \includegraphics[width=0.7\linewidth]{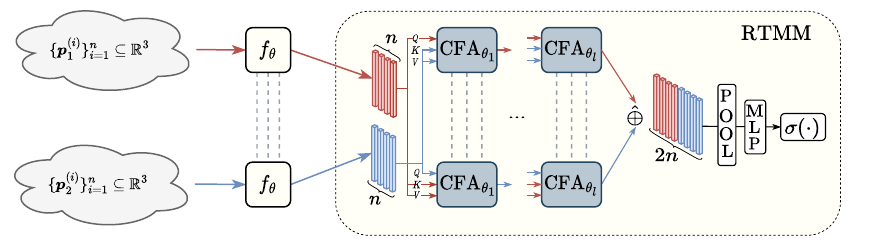}
    \caption{\textbf{The architecture of our proposed symmetric matching head, RTMM.}}
    \vspace{-13pt}
    \label{fig:model}
\end{figure*}

\subsection{3D Single Object Tracking}
Single Object Tracking (SOT) from point clouds focuses on the task of identifying a single target object within a large search area. Consequently, most methods apply point processing networks to compare the target to the search area, which can be adapted to our point cloud ReID setting. 
Indeed, many recent works~\cite{giancola2019completion,hui2022siamesepointtrans,hui2021v2b,shan2022sotwt,qi2020p2b} have shown the benefits of siamese point-processing networks for SOT. Giancola et al.~\cite{giancola2019completion} learn a similarity function between cropped point cloud patches and use a Kalman filter to generate candidate bounding boxes for matching the current target observation. 
In follow-up work, P2B~\cite{qi2020p2b} eliminates the need to approximate greedy search at inference time in favor of an end-to-end regression approach that directly estimates the target's next position through Hough voting. Hui et al. \cite{hui2021v2b} improve on this approach with a novel regression head inspired by work on object detection \cite{ge2020afdet}. 
In their latest work~\cite{hui2022siamesepointtrans}, the authors improve on their previous results by employing a point-transformer backbone. Given the strong performance of their architecture for SOT, we include it in our study and show that the point transformer also attains strong performance for object re-identification.

\subsection{Object Re-Identification from images and point clouds}
ReID from images has been an active area of research for many years, with most work focusing on Vehicle ReID~\cite{zhu2020vocreid,zheng2020beyond,sebastian2020dual,eckstein2020large,he2020multidomain,gao2020complementary,liu2020nonlocal,qian2022decoupling,zhao2021hausdorff,khorramshahi2020devil,zheng2021vehiclenet} or Person ReID~\cite{zheng2019joint,zheng2019indentity,zheng2018cnn,hermans2017defense,luo2019tricks,wang2018granularity,zeng2020hardbatch,he2021transreid,zheng2019consistent,chung2017stream,hu2021hand,jiao2022dynamic,jincloth2022}. In contrast, ReID from point clouds has received relatively little attention. A number of works consider ReID from RGB-D data~\cite{munaro2014oneshot,liciotti2016topview,liu2017online,patruno2019skeleton}, leveraging image and depth information in conjunction with skeleton normalization to improve re-identification performance in classical computer vision pipelines. 
In more recent work, Zheng et al.~\cite{zheng20223dreid} propose OG-NET for pedestrian ReID from synthetic point clouds (generated from images using a human pose estimation pipeline) and RGB information. 
While we also use a deep neural network to process point clouds for re-identification, our study involves real observations of multiple different classes cropped directly from large-scale autonomous driving datasets. 



\section{Method}
In the following section, 
we detail the architecture of our proposed Real-Time Matching Module (RTMM) for making efficient pairwise comparisons between object observations; we illustrate how existing point-based architectures can be adapted to use it, leading to a family of RTMM-based point cloud ReID networks; and we define our training objective.

\subsection{A real-time matching mechanism for point sets}
\begin{figure}
    \centering
    \includegraphics[width=0.9\linewidth]{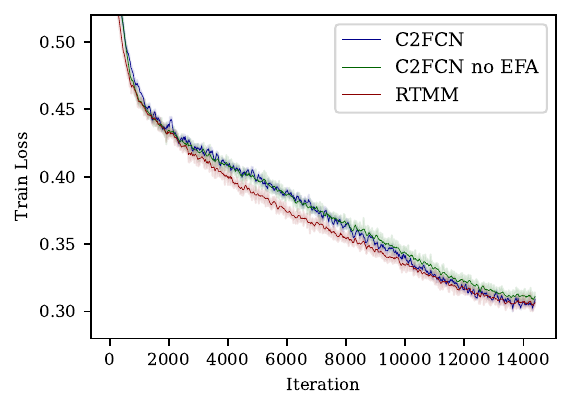}
    \vspace{-12pt}
\caption{\textbf{RTMM learns faster during training and converges to a lower training loss.} Each curve is an average over $4$ trials with different seeds. Shaded regions represent one standard error from the mean.}
\vspace{-15pt}
    \label{fig:training-loss}
\end{figure}

Given our goal of evaluating point cloud ReID in a setting that is relevant to many applications, 
it is important for our matching module to be capable of making a large number of pairwise comparisons (e.g., between tracks and detections from one time step to the next for multi-object tracking) in real-time. While performant architectures for comparing pairs of point clouds exist in the single object tracking literature, these methods lack an attunement to our real-time re-identification setting as they are too slow and consider an asymmetric search problem where the target and search area are not interchangeable. To construct an attuned architecture for re-identification without re-inventing the wheel, we select a state-of-the-art single object tracking method  \cite{hui2022siamesepointtrans} and modify their ``Coarse-to-Fine Correlation Network" (C2FCN), making it symmetric and real-time. We designate the resulting matching head RTMM (see fig.~\ref{fig:model}). RTMM achieves improved inference speed and generalization compared to the original C2FCN of \cite{hui2022siamesepointtrans}(see table.~\ref{table:head-ablation}). Moreover, RTMM's symmetric structure improves its convergence during training, allowing it to reach a lower training loss in a shorter period of time (see Figure.~\ref{fig:training-loss}).

\def\checkmark{\tikz\fill[scale=0.4](0,.35) -- (.25,0) -- (1,.7) -- (.25,.15) -- cycle;} 

\linespread{0.8} 
\renewcommand{\arraystretch}{0.8} 
\begin{table}[ht]
\centering
\fontsize{7}{12.0}\selectfont 
\begin{tabularx}{0.9\linewidth}{l|lll}
\toprule
Model& Match Acc.&Inference speed  & Par. \\\midrule
C2FCN \cite{hui2022siamesepointtrans}&$86.39\pm0.04$\%&$92\pm7.73$ms&$182.5$k\\
C2FCN no EFA&$86.19\pm 0.08$\%&${\bf6.27}\pm1.43$ms&$91.3$k\\
RTMM&${\bf86.69}\pm0.12$\%&$13.2\pm1.48$ms&$91.3$k\\
\bottomrule
\end{tabularx}
    \vspace{-7pt}
    \linespread{1} 
\renewcommand{\arraystretch}{1} 
\caption{\textbf{RTMM generalizes better than other approaches while being reasonably efficient.} We train Point-Transformer models on WOD (over $4$ seeds) with different match heads and evaluate their performance (matching accuracy $\pm$ standard error) on \emph{Waymo Eval}. Inference speed is measured for a batch of 512 examples on an RTX 3090 GPU. }
\label{table:head-ablation}
\end{table}
    \linespread{1} 
\renewcommand{\arraystretch}{1}






Starting from 
C2FCN, we improve the module's runtime by eliminating the ego feature augmentation module. We found that its memory and computational complexity scales poorly to a large number of comparisons (as is typical in our setting) and note that the module has little benefit to accuracy in our ReID setting.
With the ego feature augmentation modules removed, only Cross-Feature Augmentation (CFA) modules remain, which are essentially linear attention blocks \cite{Katharopoulos2020linearattn}. While we preserve the internal CFA block structure (see sec.~\ref{sec-appendix:cfablock}), we modify the interleaving of CFA modules to make a forward pass through RTMM symmetric with respect to each input. Specifically, to make symmetric comparisons between two sets of points $\{\vx^{(i)}_1\}_i^{n_1} , \{\vx^{(i)}_2\}_i^{n_2}$, with $\vx_1,\vx_2 \in \R^{3}$, we apply CFA blocks symmetrically to each point cloud observation's representation allowing each point cloud to play the role of key/value and query in turn:
\begin{align}
    \bar{\mX}^l_1 = \mathrm{CFA}_{\theta_l}(\bar{\mX}^{l-1}_1,\mX_1,\bar{\mX}^{l-1}_2 ,\mX_2)\\
    \bar{\mX}^l_2 = \mathrm{CFA}_{\theta_l}(\bar{\mX}^{l-1}_2 ,\mX_2,\bar{\mX}^{l-1}_1,\mX_1).
\end{align}

Where the two sets of points $\{\vx^{(i)}_1\}_i^{n_1}, \{\vx^{(i)}_2\}_i^{n_2}$, are designated in stacked matrix form $\mX_1, \mX_2 \in \R^{n \times 3}$ (a convention we follow henceforth) and $\bar{\mX}^0_i = f_\theta(\mX_i)$ with $f_\theta$ being any point processing network. After being processed by $l$ CFA blocks in our symmetric formulation, outputs are concatenated along the sequence dimension to which we apply an invariant pooling operation: $\mathrm{pool}(\bar{\mX}^l_1 \hat{\oplus} \bar{\mX}^l_2)$. This differs from the original setup of \cite{hui2022siamesepointtrans}, which only allows one point cloud to play the role of the query. Finally, an MLP is applied to the pooled representation:
\begin{align}
    \mathrm{RTMM}^l_\theta(\mX_1,\mX_2) = \mathrm{MLP}_{res}(\mathrm{pool}(\bar{\mX}^l_1 \hat{\oplus} \bar{\mX}^l_2))
\end{align}
where $\mathrm{MLP}_{res}$ is a residual MLP block followed by a linear projection layer, mapping each output to $\R$, $\mathrm{pool}(\vx):=\mathrm{maxpool}(\vx) \oplus \mathrm{avgpool}(\vx)$; $\hat{\oplus}$ designates sequence/set level concatenation; and $\oplus$ designates vector concatenation of the channel dimension. In practice, we find that setting $l=2$ is sufficient to achieve strong matching performance. We note that on all datasets and for all point models, we subsample or resample the input point cloud to contain $n=128$ points.

\linespread{0.6} 
\renewcommand{\arraystretch}{0.6} 
\begin{table*}[ht]
    \centering
    \fontsize{6.5}{12.0}\selectfont 
    \begin{tabularx}{0.99\linewidth}{lll|lll|lllllll}
    \toprule
    &Backbone & Par. &Acc. & F1 Pos.&F1 Neg.& Car & Pedestrian &Bicycle&Bus&Motorcycle&Truck&FP\\\midrule
    \multirow{7}{*}{\hspace{-7pt}\rotatebox[origin=l]{90}{$\overbrace{\hspace{50pt}}^{\text{\emph{nuScenes Eval}}}$}}&\textbf{DeiT-base}$^{*I}$  &$85.7$M  &$92.93\%$&$92.76\%$&$93.1\%$&$95.07\%$&$89.7\%$&$89.06\%$&$91.88\%$&$90.29\%$&$92.4\%$&$96.4\%$\\
    &\textbf{DeiT-tiny}$^{*I}$  &$5.7$M  &$91.94\%$&$91.89\%$&$91.99\%$&$94.09\%$&$88.23\%$&$89.16\%$&$90.6\%$&$89.92\%$&$92.34\%$&$94.01\%$\\
    &\textbf{DeiT-tiny}$^{I}$  &$5.7$M  &$88.15\%$&$88.19\%$&$88.12\%$&$90.34\%$&$84.42\%$&$85.63\%$&$86.85\%$&$85.81\%$&$88.52\%$&$89.58\%$\\[2mm]
    &\textbf{DGCNN}$^L$  &$0.6$M &$73.37\%$&$74.25\%$&$72.42\%$&$77.19\%$&$63.71\%$&$67.1\%$&$78.98\%$&$66.31\%$&$80.53\%$&$76.73\%$\\
    &\textbf{Pointnet}$^L$&$2.8$M&$74.35\%$&$74.76\%$&$73.92\%$&$77.97\%$&$65\%$&$67.38\%$&$80.4\%$&$67.24\%$&$81.74\%$&$80.1\%$\\
    &\textbf{Point-Transformer}$^L$  &$0.5$M  &$74.54\%$&$74.72\%$&$74.35\%$&$78.36\%$&$64.39\%$&$67.24\%$&$82.62\%$&$68.08\%$&$82.48\%$&$81.04\%$\\
    &\textbf{Point-Baseline}$^L$  &$0.5$M & $74.73\%$&$74.74\%$&$74.72\%$&$78.37\%$&$65.12\%$&$67.81\%$&$80.99\%$&$67.85\%$&$82.54\%$&$83.69\%$ \\[1mm]
    \midrule
    \multirow{7}{*}{\hspace{-7pt}\rotatebox[origin=l]{90}{$\overbrace{\hspace{50pt}}^{\text{\emph{Waymo Eval}}}$}}&\textbf{DeiT-base}$^{*I}$  &$85.7$M  &$96.02\%$&$96\%$&$96.04\%$&$96.84\%$&$94.32\%$&$95.16\%$&$94.89\%$&$91.46\%$&$95.66\%$&$97.47\%$\\
    &\textbf{DeiT-tiny}$^{*I}$  &$5.7$M  & $95.42\%$&$95.43\%$&$95.4\%$&$96.3\%$&$93.69\%$&$93.06\%$&$92.91\%$&$92.43\%$&$93.04\%$&$96.33\%$\\
    &\textbf{DeiT-tiny}$^{I}$  &$5.7$M & $93.22\%$&$93.29\%$&$93.16\%$&$94.14\%$&$91.38\%$&$92.25\%$&$89.92\%$&$90.63\%$&$91.21\%$&$93.56\%$\\[2mm]
    &\textbf{DGCNN}$^L$  &$0.6$M &$84.92\%$&$85.03\%$&$84.81\%$&$86.86\%$&$80.56\%$&$84.12\%$&$80.99\%$&$74.69\%$&$89.47\%$&$90.3\%$\\
    &\textbf{Pointnet}$^L$&$2.8$M&$83.41\%$&$83.62\%$&$83.2\%$&$85.51\%$&$78.77\%$&$82.32\%$&$80.31\%$&$74.4\%$&$85.73\%$&$88.51\%$\\
    &\textbf{Point-Transformer}$^L$  &$0.5$M  &$86.99\%$&$87.16\%$&$86.81\%$&$88.84\%$&$82.94\%$&$83.89\%$&$86.46\%$&$78.92\%$&$88.58\%$&$91.51\%$\\
    &\textbf{Point-Baseline}$^L$ &$0.5$M  &$86.09\%$&$86.37\%$&$85.79\%$&$87.93\%$&$82.15\%$&$82\%$&$84.33\%$&$78.3\%$&$86.36\%$&$92.16\%$\\[1mm]
    \midrule
    \multirow{7}{*}{\hspace{-7pt}\rotatebox[origin=l]{90}{$\overbrace{\hspace{50pt}}^{\text{\emph{Waymo Eval All}}}$}}&\textbf{DeiT-base}$^{*I}$  &$85.7$M  &$83.1\%$&$83.36\%$&$82.82\%$&$83.43\%$&$82.67\%$&$80.47\%$&$82.14\%$&$76.44\%$&$81.5\%$&$83.84\%$\\
    &\textbf{DeiT-tiny}$^{*I}$  &$5.7$M  & $81.35\%$&$82.32\%$&$80.27\%$&$81.35\%$&$81.58\%$&$79.84\%$&$81.14\%$&$74.76\%$&$79.83\%$&$78.35\%$\\
    &\textbf{DeiT-tiny}$^{I}$  &$5.7$M & $78.71\%$&$80.49\%$&$76.58\%$&$78.55\%$&$79.33\%$&$77.19\%$&$76.48\%$&$71.84\%$&$76.91\%$&$72.3\%$\\[2mm]
    &\textbf{DGCNN}$^L$  &$0.6$M&$83.15\%$&$83.24\%$&$83.05\%$&$85.49\%$&$78.22\%$&$82.26\%$&$80.46\%$&$74.05\%$&$86.46\%$&$88.41\%$\\
    &\textbf{Pointnet}$^L$&$2.8$M&$81.62\%$&$81.88\%$&$81.35\%$&$84.02\%$&$76.62\%$&$81.37\%$&$78.16\%$&$72.1\%$&$83.9\%$&$85.76\%$\\
    &\textbf{Point-Transformer}$^L$  &$0.5$M  &$85.01\%$&$85.15\%$&$84.87\%$&$87.16\%$&$80.53\%$&$81.68\%$&$83.07\%$&$78.64\%$&$87.1\%$&$89.36\%$\\
    &\textbf{Point-Baseline}$^L$ &$0.5$M  &$84.3\%$&$84.48\%$&$84.1\%$&$86.29\%$&$80.27\%$&$79.16\%$&$80.71\%$&$77.35\%$&$86.03\%$&$90.39\%$\\
    \bottomrule
    \end{tabularx}
    \vspace{-8pt}
     \begin{flushleft}\hspace{8pt}$^*$: Pre-trained \& fine-tuned on ImageNet 1k, $^I$: using RGB data, $^L$: using LiDAR data\end{flushleft}
     \vspace{-12pt}
    \caption{\textbf{Image v.s. point cloud performance for object re-identification.} While image models outperform their point-based counterparts, the large performance improvement from nuScenes to Waymo shows that increasing LiDAR sensor resolution can lead to significant performance improvement. Pedestrians benefit the most from this increase in sensor resolution, showing that at higher resolutions even the re-identification of deformable objects is possible without any explicit skeleton normalization step. These findings show a promising future for point cloud based re-identification as LiDAR sensor resolution continues to increase.}
    
    \vspace{-10pt}
    \label{table:results}
\end{table*}
\renewcommand{\arraystretch}{1} 
\linespread{1} 

\subsection{Compatibility with existing point backbones}

In our empirical evaluation, we select $f_\theta$ to be PointNet~\cite{qipointnet2017}, DGCNN~\cite{wang2019dgcnn}, and Point Transformer~\cite{hui2022siamesepointtrans}. However, almost any point processing backbone can be adapted with minimal effort to use our proposed RTMM. Due to the unstructured nature of point cloud inputs, most point-processing backbones compute an intermediate representation $f_\theta(\mX) \in \R^{B \times N \times d}$ which is equivariant to permutations of the columns of $\mX$, followed by an invariant pooling layer. Such constructions preserve the set cardinality dimension $N$ until the pooling operation, making them amenable to processing using sequence models, such as our RTMM. Therefore, many existing point backbones can be adapted to our method by extracting their representation before invariant pooling layers. 

\subsection{Training objective}
We train our networks for object re-identification tasks using binary cross-entropy,
\begin{equation}
    \mathcal{L}(\vx,\vy) = \frac{1}{n}\sum^n_{i=1}(\vy_i\cdot\log(\vx_i)+(1-\vy_i)\log(1-\vx_i)). \label{eq:bce}
\end{equation}


\label{sec:pointbackbone}

\vspace{-5pt}
\section{Large scale point cloud ReID Datasets}
\label{sec:datasets}
To train our point re-identification networks, we extract object observations from the nuScenes dataset~\cite{ceasar2020nuscenes} and the Waymo Open Dataset (WOD)~\cite{sun2020waymo}. This extraction process is non-trivial and seeks to maximize the applicability of our results to downstream applications, such as multi-object tracking, that identify objects using an object detector as a first step. Here, we briefly describe the salient details. 
\paragraph{Sensors} Each dataset contains multimodal driving data captured from one (nuScenes) or multiple (WOD) LiDAR sensors and an array of camera sensors. NuScenes employs a single $360^\circ$ $32$-beam LiDAR, while Waymo features one $64$-beam $360^\circ$ $10$\,Hz top-mounted sensor with four additional close proximity LiDAR sensors on the front, back, and sides of the vehicle. This means that the WOD LiDAR scans will be many times denser than their nuScenes counterparts. 
The situation is reversed for cameras, however. In nuScenes, there are 6 cameras that capture a full $360^\circ$ view of the scene, while the WOD only has 5 cameras with a front-facing FOV of $\sim252^\circ$ and a corresponding blind spot behind the vehicle. These differences allow us to examine the effect of sensor resolution on ReID performance and to explore a practically relevant setting where point-based ReID trivially complements image ReID due to the camera's blind spot.
\vspace{-10pt}
\paragraph{Object Extraction} To simulate the noise encountered in a real tracking-by-detection setting we extract object observations using bounding boxes predicted by 3D object detectors. For nuScenes, we use a pre-trained BEVfusion C+L model~\cite{liu2022bevfusion}, while we train our own CenterPoint model~\cite{yin2021centerpoint} for 3D object detection on WOD (see sec.~\ref{appendix-sec:detectors} for details). 
We post-process detections by using each model's implementation of non-maximal suppression with default settings and further eliminate noisy detections by thresholding their confidence score to be above $\tau_c=0.1$. Using the remaining detections, we extract true and false positives by matching detected bounding boxes to ground truth bounding boxes using a permissive 3D Intersection over Union (IoU) threshold of $\tau_{IoU}=0.01$. 
Hungarian matching is used here to obtain a unique assignment between ground truth and true positive bounding boxes. Duplicate true positives are discarded. 
To extract observations from LiDAR scans, we first crop points within an object's 3D bounding box before translating and rotating them such that the 3D bounding box becomes centered at $(0,0,0)^\top$ and faces a canonical orientation. Note that despite this normalization step, the observations will still contain realistic noise from the object detector's prediction; that is, the object's true orientation will not necessarily be facing the canonical orientation, nor will its true center necessarily be at $(0,0,0)^\top$. To extract observations from images, we project the predicted 3D bounding boxes to the image plane. Depending on its relative orientation,
 the bounding box may project to a non-rectangular shape. Therefore, we always use the smallest axis-aligned bounding box enclosing the projected shape. For bounding boxes that project to multiple images, we select the projection with the largest enclosing bounding box. To maintain object identities for re-identification we use the ground truth tracking labels.
\paragraph{Enhancing WOD Class labels} The nuScenes dataset provides a large number of class labels for their tracking dataset: car, bus, pedestrian, truck, bicycle, motorcycle, and trailer. WOD, however, provided substantially fewer tracking labels with their original dataset release: vehicle, pedestrian, and bicycle. In an effort to make the results between the two datasets more comparable, we enhance the WOD labels using their point cloud segmentation labels (released in a subsequent update to the dataset). Specifically, we annotate the objects within segmentation-annotated frames using majority voting of annotated points within their bounding boxes. Then, using the tracking labels, we propagate the new class of the object to all frames. This procedure expands the labels to include truck, bus, and motorcycle (see Fig.~\ref{fig:waymo-pie}).
\paragraph{Sampling at training time}  At training time, one epoch constitutes one pass over every unique object in the dataset. 
For each object $O$ (let $c$ denote the class of $O$), we flip a coin to determine whether to sample a positive or negative pair. Positive pairs are created by sampling a second observation uniformly at random from the other observations of $O$, while choosing to sample a negative pair leads to another coin toss. This time, we select between sampling a false positive $\text{FP}$ of class $c'$ ($c'$ denotes a false positive misclassified by the object detector as belonging to class $c$) or a true positive by sampling an object $O'$ of class $c$ other than $O$. In either case, we must account for point density before sampling our observation. If we were to naïvely select uniformly at random among all possible observations to create a negative pair, the distributions of point densities would be wildly different between positive and negative pairs.
Intuitively, this happens because positive pairs are always sampled among observations of the same object that may be more likely to have similar numbers of points. If sampling is done naïvely, models can fit the spurious correlation created between positive samples and point density during training. To avoid this pitfall, when sampling a false or true positive observation to create a negative pair, we follow $O's$ categorical distribution over the point density buckets $[2^n,2^{n+1})$ to select the bucket from which we then sample observations. This way, the positive and negative examples follow roughly the same point density distribution during training. Table \ref{table:sampling-ablation} shows how this simple sampling algorithm which we can "Even Sampling" improves over naïvely sampling uniformly at random.  We additionally provide pseudocode for our training-time sampling procedure is detailed in algorithm~\ref{algo:sampling}.

\definecolor{lightgreen}{rgb}{0.56, 0.93, 0.56}

\linespread{0.8} 
\renewcommand{\arraystretch}{0.8} 
\begin{table}[ht]
\centering
\fontsize{7}{12.0}\selectfont 
\begin{tabularx}{0.9\linewidth}{l|llcc}
\toprule
Model& Uniform & Even & $\Delta$& Eval Dataset \\\midrule
DGCNN&$71.6$\%&$73.37$\%&\cellcolor{lightgreen}+1.77&\multirow{3}{*}{\emph{Waymo Eval All}}\\
PointNet&$72.92$\%&$74.35$\%&\cellcolor{lightgreen}+1.43&\\
Point-Transformer&$72.83$\%&$74.54$\%&\cellcolor{lightgreen}+1.71&\\ \midrule
DGCNN&$82.26$\%&$84.92$\%&\cellcolor{lightgreen}+2.66&\multirow{3}{*}{\emph{nuScenes Eval}}\\
PointNet&$81.37$\%&$83.41$\%&\cellcolor{lightgreen}+2.04&\\
Point-Transformer&$84.75$\%&$86.99$\%&\cellcolor{lightgreen}+2.24&\\
\bottomrule
\end{tabularx}
\vspace{-10pt}
\caption{\textbf{Even sampling improves performance for all models across both datasets.}}

\vspace{-10pt}
\label{table:sampling-ablation}
\end{table}
\linespread{1} 
\renewcommand{\arraystretch}{1} 

\paragraph{Sampling at testing time} At testing time, we  sample a balanced test set of large size that can accurately estimate the performance of our models at all point densities. To accomplish this, we sample at most $10$ distinct positive pairs ($o_1$,$o_2$) for each object in the test set, keeping track of their point densities($d_{o_1}$,$d_{o_2}$). Then, for each positive pair ($o_1$,$o_2$), we sample a corresponding negative pair ($o_1$,$o_2'$), where $o_2'$ has a similar point density to $o_2$. We define point densities as similar if they fall within the same power-two interval: $[2^n,2^{n+1})$. Before sampling from the nuScenes test set, we filter out observations that have fewer than two points and observations without image crops. 
We name this test set \emph{nuScenes Eval}. 
On WOD, we create two test sets. The first, called \emph{Waymo Eval}, is created identically to \emph{nuScenes Eval}.
The second, called \emph{Waymo Eval All}, only filters out observations that have fewer than two points. Therefore, it will include many observations that have no associated image crops as they are out of the sensor's field of view, exposing the actual performance of the image models. 
Table~\ref{table:eval-stats} reports statistics of these evaluation sets.

\begin{figure*}[ht]
    \centering
    \subfloat{{\includegraphics[width=0.49\linewidth]{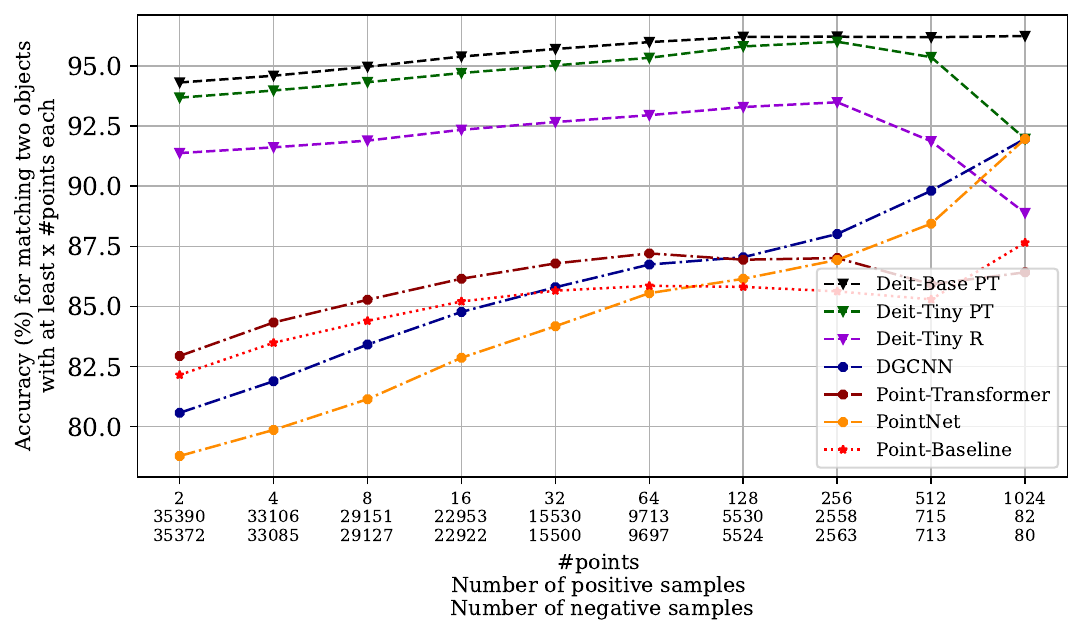} }}
    \subfloat{{\includegraphics[width=0.49\linewidth]{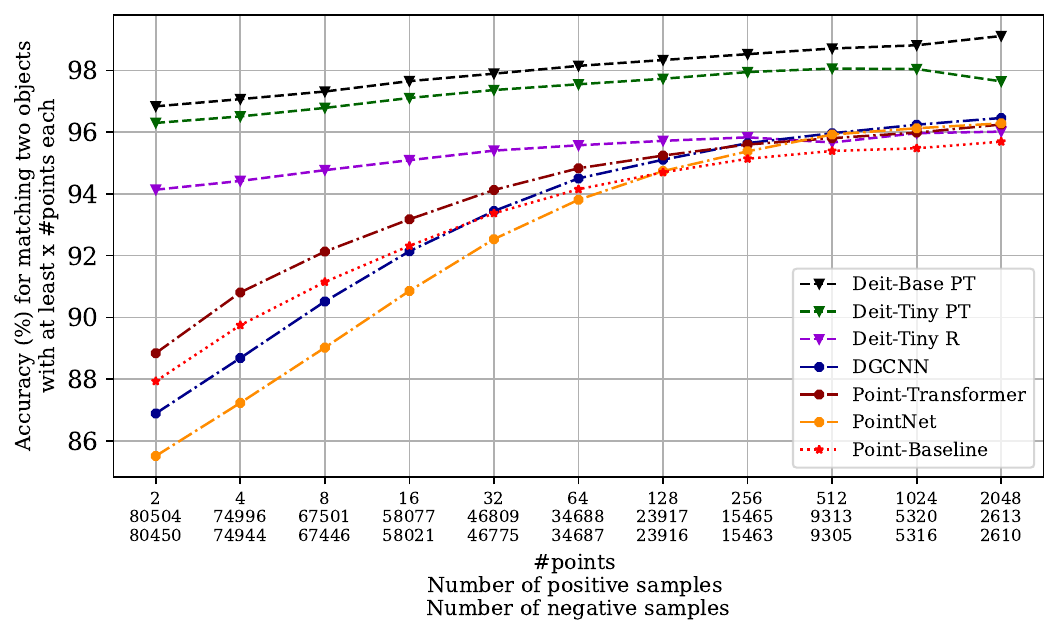} }}
    \vspace{-8pt}
    \caption{\textbf{Deformable v.s. rigid objects.} We plot the performance of models trained on WOD and evaluated on \emph{Waymo Eval} as a function of point density for class pedestrian (left) and class car (right). Performance on rigid objects is much stronger than for deformable objects. However, our results show that re-identification of deformable objects can be learned directly from data without the need of explicit skeleton normalization. }
    \vspace{-16pt}
    \label{fig:deformable}
\end{figure*}

\vspace{-5pt}
\section{Experiments}
Our empirical evaluation is based on two re-identification datasets created from nuScenes and WOD (details provided in sec.~\ref{sec:datasets}). The difference in LiDAR resolution between each dataset (32 v.s. 64 beam, respectively), allows us to establish how ReID performance varies as sensor resolution increases. We also establish the relative performance of image-based and point-based ReID, show how performance varies with respect to point density within a dataset, demonstrate that increasing compute budget significantly increases our models' performance, and fit a power-law fit to extrapolate point cloud ReID performance given more compute.



\subsection{Experimental Details}


To place our experiments within a meaningful context, we train three image models and one point cloud baseline model to compare with our three point cloud ReID networks (PointNet \cite{qipointnet2017}, DGCNN \cite{wang2019dgcnn}, and Point-Transformer \cite{hui2022siamesepointtrans}). For our image baselines, we select DeiT \cite{touvron2021deit}, a family of efficient vision transformers of different sizes. They are efficient and can be adapted with little effort to use our proposed RTMM, unlike CNNs. Specifically, we choose DeiT-Tiny as our main point of comparison and train one DeiT-Tiny model from a pre-trained checkpoint and another from random initialization. DeiT-Tiny allows us to assess the performance of an image model with a \emph{comparable} number of parameters to our point models ($5$M v.s. $2.8$M). We also train a larger DeiT-Base model from a pre-trained checkpoint for reference. To compare RTMM to another matching head as a baseline, we select C2FCN no EFA due to its real-time efficiency in combination with the point-transformer backbone.

All models were trained using identical hyperparameters and the final checkpoint is used for evaluation. We used the AdamW \cite{loshchilov2019adamw} optimizer with a learning rate of $1e-5$, weight decay of $0.01$, cosine learning rate and momentum schedules \cite{smith2019super}, and gradient clipping of 2-norm $1$. We use a batch size of $256 \times 4$ GPUs and $60 \times 4$ GPUs for point cloud and image experiments respectively and note that our batch normalization layers were not synchronized across devices during training.  Pre-trained models are trained for $200$ epochs each, while models trained from scratch are optimized for $500$ epochs and $400$ epochs on nuScenes and Waymo, respectively. The number of gradient descent steps for randomly initialized models is roughly the same ($\pm 3$ epochs) across both datasets as the Waymo dataset is larger.



\subsection{A Comparison between point-based and image-based ReID} \label{sec:pointvsimage}
Table.~\ref{table:results} reports the results of our large-scale empirical study. The top section of the table corresponds to models trained on nuScenes and evaluated on \emph{nuScenes Eval}. The bottom two sections correspond to models trained on Waymo and evaluated on \emph{Waymo Eval} and \emph{Waymo Eval All}, respectively. Matching accuracy is reported overall and for each individual class. We also report F1 scores for positive and negative matches. These results shed light on how point cloud ReID performance improves as sensor resolution is increased, how point-ReID performance varies for different objects and object categories (e.g. rigid v.s. deformable), and the performance difference between point-based and image-based object re-identification.


When comparing the accuracy of models trained on nuScenes to those trained on WOD, we observe that there is an overall increase for all models. However, the point models improve by a much greater margin than image models: as much as $12.45\%$ for the point transformer versus a $5.07\%$ increase for the randomly initialized DeiT-tiny model. We hypothesize that the performance increase of image models is due to the following reasons: 1) the image sensors are of higher resolution on WOD and 2) the WOD training set is much more diverse---it has $80\%$ more objects. This second reason is a potential confounder when assessing the extent to which the increase in point density improves point ReID performance. However, under some reasonable assumptions (see sec. \ref{sec-appendix:assumption}), the smaller relative increase for image models allows us to account for the confounding effect of a more diverse training set on WOD, showing that the increase in sensor resolution from nuScenes to WOD causes a performance improvement of at least $12.45\%-5.07\%=7.38\%$ for our point ReID models. This substantial increase in performance to $86.99\%$ accuracy from the top-performing Point-Transformer model shows that reasonable accuracy can be obtained from point-based ReID with enough sensor resolution. 

All our models on both datasets learn an unbiased matching function on aggregate as is evidenced by similar positive and negative F1 scores.  When looking at class-specific results, we note that all models follow a similar increase from nuScenes to WOD as can be observed for accuracy, except for some image models whose performance decreases on the Bus class. Of all classes, pedestrian and bicycle benefit the most from the increase in LiDAR sensor resolution with respective increases of  $18.01\%$ and $14.67\%$. This is a boon for point cloud ReID's applicability to downstream applications as it shows that the re-identification of deformable objects can be learned directly from the data when sensor resolution is sufficiently large. We note that truck and bus benefit the least from increasing LiDAR sensor resolution. We hypothesize that this is because large objects will have many points regardless of the sensor's resolution. 

Comparing the point re-identification models, Point-Transformer performs best on WOD, while all models perform very similarly on the nuScenes dataset. Focusing on image models exclusively, we note that the pre-trained DeiT-Base model performs best of all as is expected given its large number of parameters. Directly comparing point models to image models, we observe that image models always outperform their point-based counterparts when observations are visible to both camera and LiDAR sensors, but that increasing sensor resolution considerably decreases this gap. When comparing the Point-Transformer to the randomly initialized DeiT-Tiny on WOD, we observe the smallest performance gap between large rigid objects (bus, truck, and car), while the smaller deformable objects (pedestrian and bicycle) pose more difficulty to the Point-Transformer and point-based models in general. This is to be expected as deformable objects create inherent shape ambiguity, which can be resolved in images by leveraging color or texture information, but for point clouds, an object's shape is its primary distinguishing characteristic. While the point models perform poorer than image models overall, the relative improvement seen from nuScenes to WOD is non-trivial and suggests that the gap in performance will shrink as LiDAR sensor resolution continues to increase.

\begin{figure}
    \centering
    \includegraphics[width=0.9\linewidth]{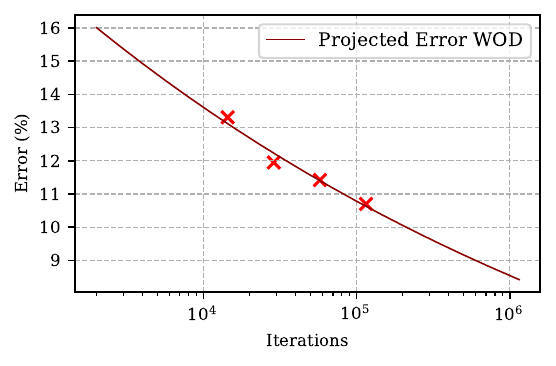}
    \vspace{-15pt}
\caption{\textbf{Extrapolating error as a function of training iterations.} We fit a power-law to our training curve, yielding $\epsilon = 34.5x^{-0.1}$.}
    \vspace{-20pt}
    \label{fig:power-law}
\end{figure}

\linespread{0.8} 
\renewcommand{\arraystretch}{0.8} 
\begin{table*}[ht]
    \centering
    \fontsize{6.5}{12.0}\selectfont 
    \begin{tabularx}{\linewidth}{lll|lll|lllllll}
    \toprule
    \hspace{-18pt}&Backbone & Epochs &Acc. & F1 Pos.&F1 Neg.& Car & Pedestrian &Bicycle&Bus&Motorcycle&Truck&FP\\\midrule
    \multirow{4}{*}{\hspace{-7pt}\rotatebox[origin=l]{90}{$\overbrace{\hspace{30pt}}^{\text{\emph{Waymo Eval}}}$}}&\textbf{Point-Transformer}$^L$  &$3200$  &$89.3\%$&$89.45\%$&$89.16\%$&$90.68\%$&$86.22\%$&$87.42\%$&$89.21\%$&$84.47\%$&$90.92\%$&$92.35\%$\\
    &\textbf{Point-Transformer}$^L$  &$1600$  &$88.58\%$&$88.75\%$&$88.41\%$&$90.21\%$&$85\%$&$85.52\%$&$87.8\%$&$82.2\%$&$90.18\%$&$92.13\%$\\
    &\textbf{Point-Transformer}$^L$  &$800$  & $88.05\%$&$88.2\%$&$87.9\%$&$89.69\%$&$84.47\%$&$85.09\%$&$86.54\%$&$81.89\%$&$89.53\%$&$92.14\%$\\
    &\textbf{Point-Transformer}$^L$  &$400$  &$86.99\%$&$87.16\%$&$86.81\%$&$88.84\%$&$82.94\%$&$83.89\%$&$86.46\%$&$78.92\%$&$88.58\%$&$91.51\%$\\
    \bottomrule
    \end{tabularx}
    \vspace{-12pt}
     \begin{flushleft}\hspace{4pt}$^L$: using LiDAR data\end{flushleft}
     \vspace{-20pt}
    \caption{\textbf{Scaling compute improves performance for all classes on WOD.}}
    \vspace{-15pt}
    \label{table:scaling}
\end{table*}
\linespread{1} 
\renewcommand{\arraystretch}{1} 
\subsection{Intra-dataset point density comparison}

It is reasonable to expect that point cloud ReID performance will continue to increase as LiDAR sensor resolution increases. Figure \ref{fig:pointreidapproaches} estimates the effect that even higher sensor resolution has on performance by plotting the accuracy of each model as a function of point density. Specifically, the accuracy (y-axis) is measured for different subsets of \emph{nuScenes Eval} (left) and \emph{Waymo Eval} (right) containing pairs of point cloud observations $(\{\vx^{(i)}_1\}_i^{n_1} , \{\vx^{(i)}_2\}_i^{n_2})$, where $x \leq \min(n_1,n_2)$. The number of positive and negative examples for each threshold is shown on the x-axis. We observe that the magnitude of the increase is much greater for point models than image models, showing that when sufficient points are available, point cloud ReID models can approach the performance of image re-identification models. This further suggests that increasing LiDAR sensor resolution will improve point cloud ReID performance.

Figure.~\ref{fig:deformable} compares the re-identification performance for deformable (pedestrian, left) and non-deformable objects (car, right) as point density is increased on \emph{Waymo Eval}. For pedestrians, DGCNN and PointNet benefit the most from higher sensor resolutions, while the Point-Baseline and Point-Transformer models (both using a Point-Transformer backbone) are unable to take advantage of the highest point densities. This difficulty seems to be unique to deformable objects, however, as the car class follows a logarithmic trend of improvement with all models achieving similar performance. We note that the Point-Transformer model equipped with our proposed RTMM bests the Point-Baseline at every point density. With as few as $64$ points per object, point-based object re-identification attains performance greater than $94\%$ for all models when objects are rigid. This shows that point-based object re-identification can be extremely competitive in such settings.


\vspace{-5pt}
\section{Scaling training compute for point-based ReID}
\vspace{-5pt}
As seen in table~\ref{table:dataset-statistics} the number of samples in our ReID datasets is combinatorially large. For WOD, there are $4.35e8$ positive samples and $3.89e19$ negative samples. To put this in perspective it would take $\sim 13,646$ epochs to sample all possible positive samples on WOD while we only train our models for $400$ epochs. To provide practical estimates of attainable performance and showcase the best performance attainable we train four Point-Transformer models for $400\cdot 2^i$ epochs with $i\in\{0,1,2,3\}$ on the WOD and fit a power-law through their validation accuracy.

Tables~\ref{table:scaling} and \ref{table:scaling-nus} show the performance of models trained on progressively larger compute budgets for WOD and nuScenes, respectively. We observe a similar effect for both datasets: performance increases across the board as the compute budget is increased. Since the largest training schedules ($3200$ epochs or $115200$ iterations) on WOD only sample a small fraction of the enormous number of possible samples, we hypothesize that performance will continue to increase with more training.

Figure.~\ref{fig:power-law} plots a power law fit to the error and training iterations from table~\ref{table:scaling}. Specifically, we follow model $\mathcal{M}_2$: $\epsilon_x - \epsilon_\infty =\beta x^c$ from \cite{alabdulmohsin2022scaling}. The best fit obtained was $\epsilon_\infty=0,\beta=34.5,c=-0.1$.  This suggests that even better ReID performance is attainable by continuing to increase compute with an order of magnitude more training iterations projected to yield a model with less than 9\% error. This is encouraging for applications of point-based ReID which typically require low error to be worthwhile. 


\vspace{-7pt}
\section{Conclusion}
\vspace{-5pt}
We have conducted the first large-scale study of object re-identifications from point cloud observations. Our findings can be summarized as follows: 1) we propose RTMM a symmetric matching head for point cloud ReID that improves generalization and convergence, 2) we establish the performance of point cloud ReID relative to image ReID, 3) we show that our point ReID networks can attain strong ReID performance, even on par with image models, as long as the compared observations are sufficiently dense, 4) we established that point ReID performance increases as LiDAR sensor resolution is increased, 5) we demonstrated the performance of point ReID models can be substantially increased by training for longer (89\%+ accuracy).

While image ReID outperforms point ReID when observations are visible to both sensors, our results show that the latter still attains strong enough performance to be useful for downstream applications. Therefore, applications can be developed that leverage this newly discovered capability. For the time being, autonomous driving systems like the WOD vehicle, which have limited camera FOV, stand to benefit the most from the added complementarity of a ReID network processing $360^\circ$ LiDAR scans. However, even vehicles equipped with cameras covering $360^\circ$ can benefit from the added redundancy of point ReID, especially in cases where the observations are sufficiently dense to be reliable. In the future, as LiDAR technology continues to advance, point ReID performance can only increase---magnifying the implications of our findings. Already today, bleeding edge LiDAR sensors feature 128 beams\cite{alphaprimelidar}, twice the resolution of WOD's top-mounted LiDAR.


Our initial study opens many directions for future work. Integrating our point cloud ReID models into downstream applications such as multi-object tracking for autonomous driving or robot grasping are logical next steps. Other directions include improving ReID performance by fusing LiDAR in the camera, using techniques such as \cite{nagrani2021attentionbottleneck,wang2020cen} could work well with our framework. Another interesting direction would be to study how additional geometric priors, such as SE(3) or SO(3) equivariance \cite{deng2021vectorneuron,fuchs2020se3} can enhance ReID from point clouds.


{\small
\bibliographystyle{ieee_fullname}
\bibliography{bib/2d-mot,
              bib/3d-mot,
              bib/point-processing,
              bib/sot,
              bib/re-id,
              bib/surveillance,
              bib/siamese,
              bib/misc-other}
}

\clearpage
\begin{appendix}
\onecolumn

\begin{center}
{
\centering \Large \textbf{Supplementary Material for}}\\
{\centering \Large\textbf{Object Re-Identification from Point Clouds}
 }
\end{center}
\section{Extended experimental details}
The following section contains further details of the models used to conduct our experiments. 

\subsection{3D object detectors} \label{appendix-sec:detectors}
We generate our ReID datasets using point cloud samples cropped from predicted 3D bounding boxes and image samples cropped from their projections onto images. Here, we describe the details of the object detectors used to predict these bounding boxes.

\begin{table}[!htb]
    \begin{minipage}{.46\linewidth}
      \centering
        \begin{adjustbox}{width=\linewidth,center}
        \begin{tabular}{lcccc}
        \toprule
        \multirow{2}{*}{\textbf{Class}} & \multicolumn{2}{c}{\textbf{Level 1}} & \multicolumn{2}{c}{\textbf{Level 2}} \\
         & mAP & mAPH & mAP & mAPH \\ \midrule
        \textbf{Car} & 68.3 & 67.8 & 60.7 & 60.2 \\
        \textbf{Truck} & 34.2 & 33.5 & 32.2 & 31.6 \\
        \textbf{Bus} & 49.0 & 48.6 & 42.6 & 42.3 \\
        \textbf{Motorcycle} & 53.1 & 52.2 & 38.0 & 37.4 \\
        \textbf{Cyclist} & 68.4 & 67.0 & 65.8 & 64.5 \\
        \textbf{Pedestrian} & 65.9 & 59.7 & 58.0 & 52.4 \\ \bottomrule
        \end{tabular}
        \end{adjustbox}
        \caption{\textbf{Performance of the CenterPoint model on the WOD with enhanced vehicle labels.} The metrics are only calculated on samples with an enhanced label.}
        \label{table:waymodetector}
    \end{minipage}%
    \begin{minipage}{.01\linewidth}
    \hspace{5mm}
    \end{minipage}
    \begin{minipage}{.46\linewidth}
      \centering
      \begin{adjustbox}{width=\linewidth,center}
        \begin{tabular}{lcccccc}
        \toprule
        \textbf{Class} & \textbf{AP} & \textbf{ATE} & \textbf{ASE} & \textbf{AOE} & \textbf{AVE} & \textbf{AAE} \\ \midrule
        \textbf{Car} & 89.2 & 0.170 & 0.148 & 0.061 & 0.274 & 0.185 \\
        \textbf{Truck} & 64.6 & 0.326 & 0.181 & 0.093 & 0.247 & 0.217 \\
        \textbf{Bus} & 75.3 & 0.338 & 0.189 & 0.069 & 0.430 & 0.274 \\
        \textbf{Trailer} & 42.5 & 0.520 & 0.201 & 0.610 & 0.214 & 0.140 \\
        \textbf{Const. Veh.} & 30.4 & 0.735 & 0.431 & 0.797 & 0.118 & 0.295 \\
        \textbf{Pedestrian} & 88.2 & 0.134 & 0.288 & 0.387 & 0.217 & 0.101 \\
        \textbf{Motorcycle} & 78.6 & 0.184 & 0.249 & 0.216 & 0.348 & 0.271 \\
        \textbf{Bicycle} & 65.1 & 0.169 & 0.257 & 0.411 & 0.190 & 0.015 \\
        \textbf{Traffic Cone} & 79.5 & 0.121 & 0.317 & - & - & - \\
        \textbf{Barrier} & 72.0 & 0.178 & 0.277 & 0.054 & - & - \\ \midrule
        \textbf{NDS: 0.714} & 68.5 & 0.288 & 0.254 & 0.300 & 0.255 & 0.187 \\ \bottomrule
        \end{tabular}
      \end{adjustbox}
      \caption{\textbf{Performance of the BEVfusion C+L model on the nuScenes validation set.} This model was used to generate the nuScenes detections for tracking and to create the nuScenes ReID dataset.}  
        \label{table:bevfusioneval}
    \end{minipage} 
\end{table}

\paragraph{nuScenes} On the nuScenes dataset, we use BEVfusion C+L \cite{liu2022bevfusion} to extract detections for constructing our ReID dataset. Table~\ref{table:bevfusioneval} shows its performance on  the nuScenes validation set.

\begin{figure}[ht]
    \centering
    \includegraphics[width=0.6\linewidth]{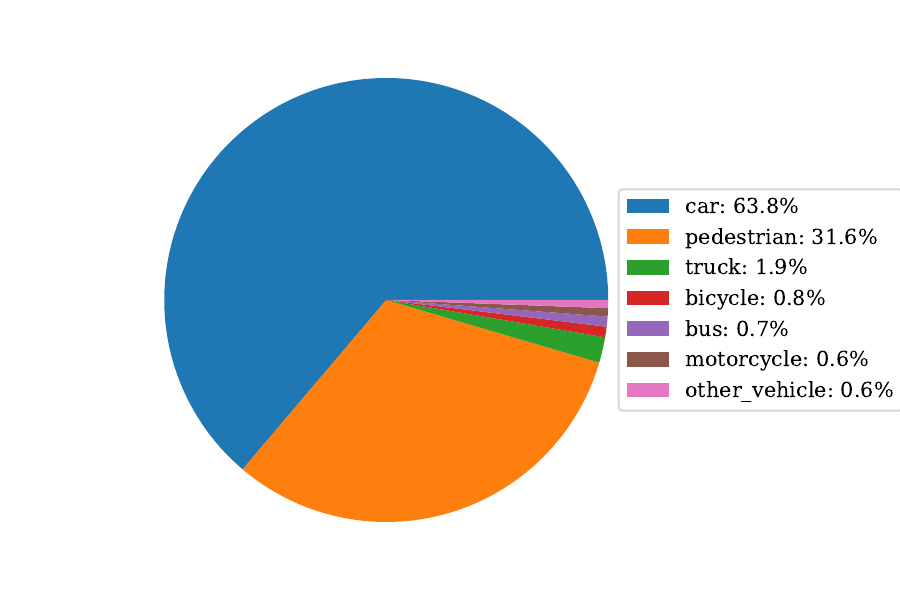}
    \caption{\textbf{Waymo class label split after propagating segmentation annotations.} The plot shows the proportions of different types of objects in the Waymo training set. We note that those samples belonging to the ``other\_vehicle" class are ignored during the creation of our ReID dataset since we do not have label information for them.}
    \label{fig:waymo-pie}
\end{figure}

\paragraph{Waymo} On the Waymo dataset, we train a CenterPoint~\cite{yin2021centerpoint} object detector using the implementation from \cite{mmdet3d2020}. Our CenterPoint model is trained for 20 epochs, where each epoch constitutes a pass over 20\% of the frames in WOD (sampled uniformly at random). Since the Waymo dataset's Vehicle class is heterogeneous and contains various sub-types of objects that are annotated separately in the nuScenes~\cite{ceasar2020nuscenes} dataset, we utilize the point-wise semantic segmentation label to automatically divide the Vehicle class into \{Car, Truck, Bus, Motorcycle\}. Figure~\ref{fig:waymo-pie} shows the proportion of object classes after propagating segmentation labels. Our model's evaluation results on the Waymo validation set with segmentation labels are reported in table~\ref{table:waymodetector}.

While results between datasets are difficult to compare, we hypothesize that the nuScenes model attains relatively stronger performance. This is to be expected as the model incorporates both camera and LiDAR information, while our Waymo model does not. We note that this performance difference should have an impact on the level of noise in predicted bounding boxes and consequently may impact our ReID network's performance.

\subsection{Hyperparameters, training details, and inference speed.}
\label{sec:alltraindetails}
Table~\ref{table:hyperparameters} reports the hyperparameters of our ReID networks in greater detail. We note that the difference in training epochs ($400$ vs. $500$) for the Waymo and nuScenes datasets was intentional to have approximately the same number of gradient descent steps for each ($\pm 3$ epochs).

\begin{table}[ht]
    \centering
    \fontsize{7}{12.0}\selectfont 
    \begin{tabularx}{0.375\linewidth}{lll}
    \toprule
    Parameter & Explanation& Value\\\midrule
    \multicolumn{3}{l}{\textbf{Point model hyperparameters}}\\
    B & Batch Size & $256 \times 4$\\
    N & Number of input points & $128$\\
    $D_p$ & Point input dimension & $3$\\
    $\mathrm{LR}$ & Learning Rate & $3\cdot10^{-4}$ \\
    WD & weight decay & $0.01$ \\
    $\mathrm{G}_c$ & Gradient Clipping & 1\\
    \multicolumn{3}{l}{\textbf{Image model hyperparameters}}\\
    $D_I$ & Image input dimension & $3\times224\times224$\\
    B & Batch Size & $60 \times 4$\\
    $\mathrm{LR}$ & Learning Rate & $3\cdot10^{-5}$ \\
    WD & weight decay & $0.01$ \\
    $\mathrm{G}_c$ & Gradient Clipping & 1\\
     \multicolumn{3}{l}{\textbf{Randomly initialized Waymo models}}\\
    E & Training epochs & 500\\
    \multicolumn{3}{l}{\textbf{Randomly initialized nuScenes models}}\\
    E & Training epochs & 400\\
    \multicolumn{3}{l}{\textbf{Pretrained Models}}\\
    E & Training epochs & 200\\
    \bottomrule
    \end{tabularx}
    \vspace{2pt}
    \caption{\textbf{Hyperparameters of our different architectures.} Both image and point models follow a cyclic learning rate schedule with target\_ratio $(10, 1e-4)$, cyclic\_times $1$, and step\_ratio\_up $0.4$ implemented by the software package MMCV \cite{mmcv}}
    \label{table:hyperparameters}
\end{table}

\linespread{0.8} 
\renewcommand{\arraystretch}{0.8} 
\begin{table}[ht]
    \centering
    \fontsize{7}{12.0}\selectfont 
    \begin{tabularx}{0.47\linewidth}{llll}
    \toprule
    Model & Par.& Batch Size & Inference Time \\\midrule
    \text{DeiT-Tiny}  &$5910$K & $100$ & $32.4\text{ ms }  \pm 1.75 \text{ms}$ \\
    \text{DeiT-Base}  &$87,338$K & $100$ & $173\text{ ms }  \pm 11.8 \text{ms}$ \\[1mm]
    \text{Point Transformer}  &$529$K & $100$ & $15\text{ ms }  \pm 0.605 \text{ms}$ \\
    \text{DGCNN}  &$617$K &  $100$ &$18.7\text{ ms } \pm 3.25\text{ ms }$\\
    \text{PointNet}  &$2800$K &  $100$ &$8.04\text{ ms }  \pm 0.0996 \text{ ms}$\\\midrule
    C2FCN\cite{hui2022siamesepointtrans}&$182.5$k&$512$&$92\pm7.73$\text{ ms }\\
    C2FCN no EFA&$91.3$k&$512$&$6.27\pm1.43$\text{ ms }\\
    RTMM&$91.3$k&$512$&$13.2\pm1.48$\text{ ms }\\
    \bottomrule
    \end{tabularx}

    \linespread{1} 
\renewcommand{\arraystretch}{1} 
    \vspace{2pt}
    \caption{\textbf{Inference speed of different point and image backbones.} All models were tested on a single RTX 3090 GPU.}
    \label{table:inference}
\end{table}

Table~\ref{table:inference} reports the inference speed of all models used in our empirical evaluation. We select batch sizes to simulate the needs of practitioners in a multi-object tracking context. Backbone models are tested for a batch size of $100$, an approximate upper bound on the number of detections a 3D object detector might make at a given timestep. We test RTMM with a larger batch size of $512$, since pairwise comparisons scale quadratically with the input size. We note, however, that it is possible to considerably reduce the comparisons needed by only comparing within the same class or imposing other such rules. At the batch sizes shown, all our models, backbone + RTMM, run in real-time ($< 10$\,Hz), except for the DeiT-Base model.

\section{Additional dataset details}
Table~\ref{table:dataset-statistics} reports training set statistics for both our ReID datasets, while table~\ref{table:eval-stats} report test set statistics. Algorithms~\ref{algo:sampling} and \ref{algo:sampling-uniform} provide pseudocode for the even and uniform sampling procedures introduces in section~\ref{sec:datasets} and evaluated in table~\ref{table:sampling-ablation}. In algorithm~\ref{algo:sampling}, for a given object, the frequency distribution over power-two buckets is computed by calculating the number of observations that fall within each bucket and normalizing by the total number of observations of that object, creating a probability distribution over buckets. 

\begin{table}[ht]
    \centering
    \fontsize{7}{12.0}\selectfont 
    \begin{tabularx}{0.4\linewidth}{llll}
    \toprule
    Dataset & Positive Pairs & Negative Pairs& Total\\\midrule
    \emph{nuScenes Eval} &53,787&53,733&107,520\\
    \emph{Waymo Eval} &120,859&120,805&241,664\\
    \emph{Waymo Eval All} &145,458&145,358&290,816\\
    \bottomrule
    \end{tabularx}
    \vspace{2pt}
    \caption{\textbf{Evaluation set statistics.} }
    \label{table:eval-stats}
\end{table}

\begin{table*}[!htb]
    \begin{minipage}{.5\linewidth}
      \centering
      \begin{adjustbox}{width=\linewidth,center}
        \begin{tabular}{lrrrr}
        \toprule
        Classes &\# of Obj.&\# of Obs.& Pos. Pairs & Neg. Pairs\\ \midrule
        FP bicycle&$--$&$13,100$&$--$&$--$\\
        FP bus&$--$&$2,069$&$--$&$--$\\
        FP car&$--$&$125,223$&$--$&$--$\\
        FP motorcycle&$--$&$7,222$&$--$&$--$\\
        FP pedestrian&$--$&$106,206$&$--$&$--$\\
        FP trailer&$--$&$8,023$&$--$&$--$\\
        FP truck&$--$&$21,792$&$--$&$--$\\
        bicycle&$554$&$7,575$&$7.76e+4$&$1.62e+12$\\
        bus&$422$&$8,375$&$1.07e+5$&$4.54e+11$\\
        car&$20,830$&$326,967$&$3.77e+6$&$3.34e+16$\\
        motorcycle&$588$&$8,201$&$8.67e+4$&$9.73e+11$\\
        pedestrian&$9,112$&$156,852$&$1.83e+6$&$5.43e+15$\\
        trailer&$773$&$13,921$&$1.66e+5$&$3.34e+12$\\
        truck&$2,933$&$50,487$&$5.98e+5$&$1.32e+14$\\\midrule
        Total&$35,212$&$856,013$&$6.63e+06$&$3.90e+16$\\
        \bottomrule
        \end{tabular}
      \end{adjustbox}
    \end{minipage} 
    \begin{minipage}{.5\linewidth}
      \centering
        \begin{adjustbox}{width=\linewidth,center}
        \begin{tabular}{lrrrr}
        \toprule
        Classes &\# of Obj.&\# of Obs.& Pos. Pairs & Neg. Pairs\\ \midrule
        FP bicycle&$--$&$27,043$&$--$&$--$\\
        FP bus&$--$&$2,378$&$--$&$--$\\
        FP car&$--$&$24,971$&$--$&$--$\\
        FP motorcycle&$--$&$46,111$&$--$&$--$\\
        FP pedestrian&$--$&$604,021$&$--$&$--$\\
        FP truck&$--$&$5,987$&$--$&$--$\\
        bicycle&$497$&$46,204$&$3.02e+6$&$1.23e+14$\\
        bus&$378$&$43,639$&$3.35e+6$&$4.59e+13$\\
        car&$43,305$&$4,002,934$&$2.78e+8$&$3.25e+19$\\
        motorcycle&$355$&$32,317$&$2.18e+6$&$9.90e+13$\\
        pedestrian&$18,107$&$1,954,970$&$1.41e+8$&$6.40e+18$\\
        truck&$1,114$&$108,722$&$7.77e+6$&$7.13e+14$\\\midrule
        Total&$63,756$&$6,899,297$&$4.35e+08$&$3.89e+19$\\
        \bottomrule
        \end{tabular}
        \end{adjustbox}
    \end{minipage}%
    \caption{\textbf{Training set statistics for nuScenes (left) and WOD (right).} From left to right, the columns contain the number of unique objects of each class, the number of observations of these objects, the number of positive pairs of each class, and the number of negative pairs of each class. False positives are used to create negative pairs with observations of the corresponding true positive class. Positive and negative pairs are created from objects of the same predicted class.} 
    \label{table:dataset-statistics}
\end{table*}


\begin{minipage}{.46\textwidth}
\centering
\begin{algorithm}[H]
\scriptsize
\SetAlgoLined
\KwData{Dataset $D$ of all objects and their observations.}
\KwResult{A set of sampled pairs for one epoch of training.}
\BlankLine
\tcp{Initialize sample}
$S \gets \emptyset$;
\BlankLine
\ForEach{object $O$ in $D$}{
$o_1 \gets \text{random observation of } O$\;
$c \gets \text{class of } O$\;
$p_1 \gets \text{random number between } 0 \text{ and } 1$\;
\If{$p_1 \leq 0.5$}
{
    \tcp{Sample positive pair} 
    $o_2 \gets \text{random observation of } O \text{ other than } o_1$\;
}
\Else
{
    \tcp{Get distribution of object O}
    $D \gets \text{frequency distribution of power-two buckets}$\\
    \hspace{20pt} for O's observations;\\
    $b \gets \text{sample a point density bucket from $D$}$;\\[1mm]
    \tcp{Sample negative observation}
    $p_2 \gets \text{random number between } 0 \text{ and } 1$\;
    \If{$p_2 \leq 0.5$}
    {
        \tcp{Sample false positive}
        $fp_o \gets \text{random false positive of density $b$ with predicted class } c$\;
        $S \gets S \cup {fp_o}$\;
    }
    \Else{
        \tcp{Sample true positive}
        $o_2 \gets \text{random observation of density b with class } c$ \\
         \hspace{20pt}$\text{ other than those associated with } O$\;
    }
}
$S \gets S \cup (o_1,o_2)$\;
}
\BlankLine
\KwRet{$S$}
\caption{Even Data Sampling Algorithm for one Epoch of Training}
\label{algo:sampling}
\end{algorithm}
\end{minipage}\hfill
\begin{minipage}{.46\textwidth}
\centering
\begin{algorithm}[H]
\scriptsize
\SetAlgoLined
\KwData{Dataset $D$ of all objects and their observations.}
\KwResult{A set of sampled pairs for one epoch of training.}
\BlankLine
\tcp{Initialize sample}
$S \gets \emptyset$;
\BlankLine
\ForEach{object $O$ in $D$}{
$o_1 \gets \text{random observation of } O$\;
$c \gets \text{class of } O$\;
$p_1 \gets \text{random number between } 0 \text{ and } 1$\;
\If{$p_1 \leq 0.5$}{
\tcp{Sample positive pair} 
$o_2 \gets \text{random observation of } O \text{ other than } o_1$\;
}
\Else{
\tcp{Sample negative observation}
$p_2 \gets \text{random number between } 0 \text{ and } 1$\;
\If{$p_2 \leq 0.5$}{
\tcp{Sample false positive}
$fp_o \gets \text{random false positive of predicted class } c$\;
$S \gets S \cup {fp_o}$\;
}
\Else{
\tcp{Sample true positive}
$o_2 \gets \text{random observation of class } c$ \\
 \hspace{20pt}$\text{ other than those associated with } O$\;
}
}
$S \gets S \cup (o_1,o_2)$\;
}
\BlankLine
\KwRet{$S$}
\caption{Uniform Data Sampling Algorithm for one Epoch of Training}
\label{algo:sampling-uniform}
\end{algorithm}
\end{minipage}


\section{A reasonable assumption}\label{sec-appendix:assumption}
Section~\ref{sec:pointvsimage} analyses the results from table~\ref{table:results} and claims that the performance improvements from nuScenes to Waymo are due in part to the increase in sensor resolution between the datasets. However, Waymo is also much more diverse than nuScenes featuring $63,756$ unique objects v.s. $35,212$. This is a potential confounder since a more diverse dataset can also lead to improved performance. We claim, in the main manuscript, that under a reasonable assumption, we can obtain a lower bound on the performance improvement from nuScenes to Waymo that can be attributed to the increase in sensor resolution. The assumption is that the increased diversity from nuScenes to Waymo causes the same performance improvement (in terms of matching accuracy) for image and point cloud models alike. We believe it is reasonable to assume this since we have paired data (i.e., the point clouds and image crops are observations of the same objects). Assuming this, we can then take the performance difference from nuScenes to Waymo of the randomly initialized DeiT-Tiny model to be the increase caused by dataset diversity (as it can only be smaller than or equal to this quantity). Then, assuming no other confounders are present, we can account for the effect of a more diverse dataset by subtracting the performance improvement of the randomly initialized DeiT-Tiny model from the improvement of the point models. This is the number we report in section~\ref{sec:pointvsimage}.

\section{Additional results analysis}
While the main manuscript highlights the results that are central to our contributions, some secondary figures could not be included, but still provide additional infromation. We include them and a corresponding discussion here. 

\subsection{Scaling compute on nuScenes}
Table~\ref{table:scaling-nus} reports results for scaling compute on nuScenes. Specifically, we train four Point-Transformer models for $500 \cdot 2^i$ epochs with $i \in \{0, 1, 2, 3\}$. All models noticeably improve from longer training, as they do on WOD.

\linespread{0.8} 
\renewcommand{\arraystretch}{0.8} 
\begin{table*}[h]
    \centering
    \fontsize{6.5}{12.0}\selectfont 
    \begin{tabularx}{\linewidth}{lll|lll|lllllll}
    \toprule
    \hspace{-18pt}&Backbone & Epochs &Acc. & F1 Pos.&F1 Neg.& Car & Pedestrian &Bicycle&Bus&Motorcycle&Truck&FP\\\midrule
    \multirow{4}{*}{\hspace{-7pt}\rotatebox[origin=l]{90}{$\overbrace{\hspace{30pt}}^{\text{\emph{nuScenes Eval}}}$}}&\textbf{Point-Transformer}$^L$  &$4000$  &$77.89\%$&$77.92\%$&$77.85\%$&$82.03\%$&$67.63\%$&$71.82\%$&$84.15\%$&$71.35\%$&$84.77\%$&$83.18\%$\\
    &\textbf{Point-Transformer}$^L$  &$2000$  &$76.91\%$&$77.09\%$&$76.72\%$&$81.02\%$&$66.68\%$&$70.25\%$&$83.01\%$&$71.3\%$&$83.78\%$&$82.21\%$\\
    &\textbf{Point-Transformer}$^L$  &$1000$  &$75.67\%$&$75.92\%$&$75.41\%$&$79.45\%$&$65.79\%$&$69.82\%$&$82.27\%$&$68.97\%$&$82.94\%$&$81.43\%$\\
    &\textbf{Point-Transformer}$^L$  &$500$  &$74.54\%$&$74.72\%$&$74.35\%$&$78.36\%$&$64.39\%$&$67.24\%$&$82.62\%$&$68.08\%$&$82.48\%$&$81.04\%$\\ 
    \bottomrule
    \end{tabularx}
    \vspace{-12pt}
     \begin{flushleft}\hspace{4pt}$^L$: using LiDAR data\end{flushleft}
     \vspace{-20pt}
    \caption{\textbf{Scaling compute improves performance for all classes on nuScenes.}}
    \vspace{-10pt}
    \label{table:scaling-nus}
\end{table*}
\linespread{1} 
\renewcommand{\arraystretch}{1} 

\subsection{Additional analysis comparing accuracy at different point densities}
\begin{figure*}[ht]
    \centering
    \subfloat{{\includegraphics[width=0.49\linewidth]{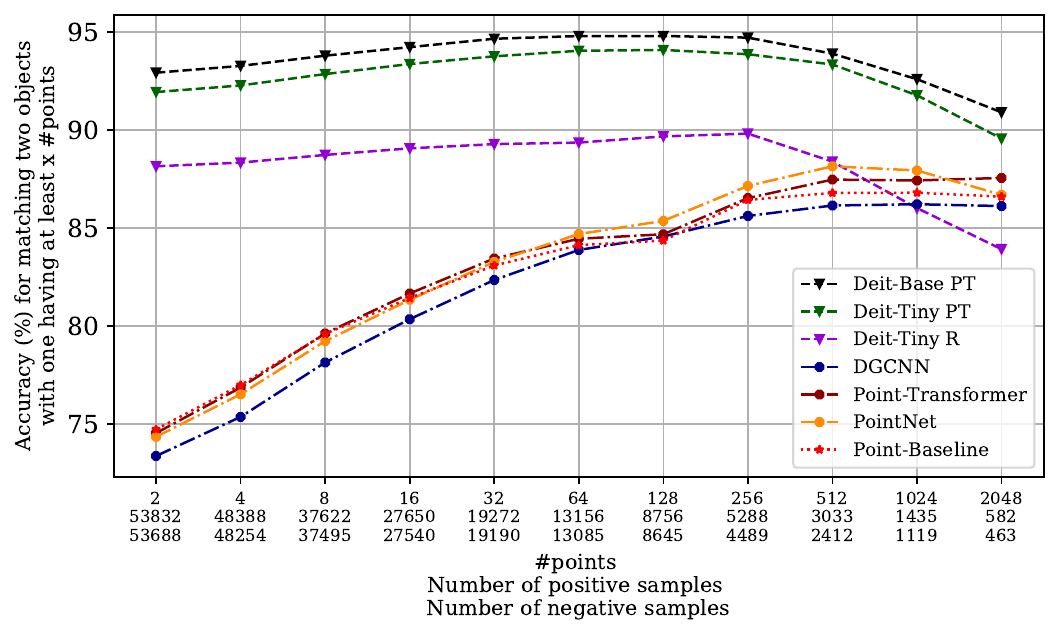} }}
    \subfloat{{\includegraphics[width=0.49\linewidth]{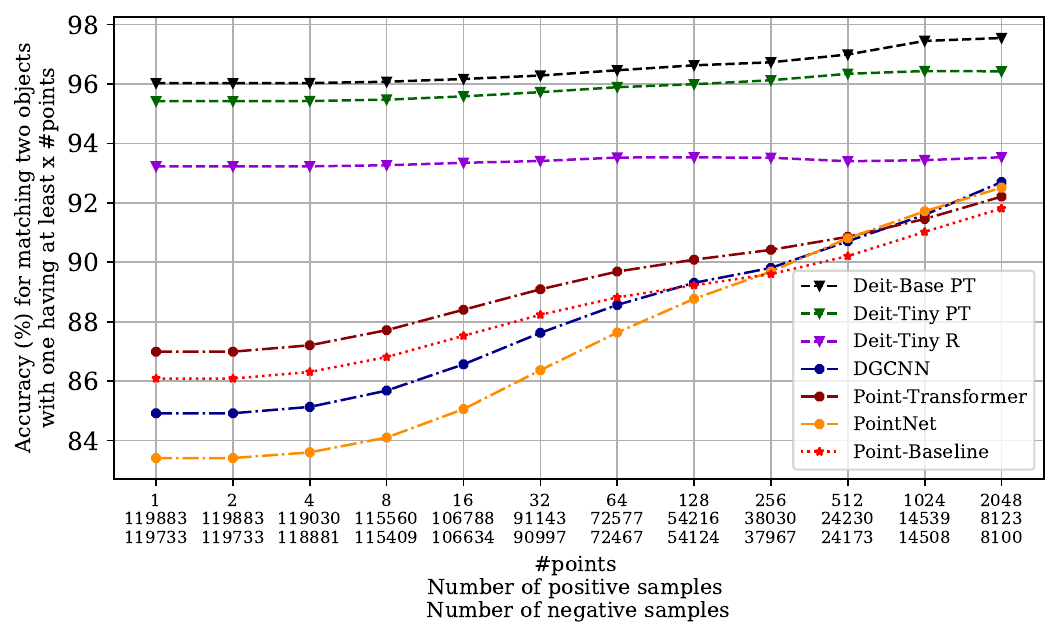} }}
    \caption{\textbf{The performance of point cloud ReID approaches image ReID with sufficient points.} The plot shows the performance of the image and point cloud ReID networks as a function of the number of points in at least one observation of the pair. Left plots models trained on nuScenes and evaluated on the \emph{nuScenes Eval} set, while right plots models trained on Waymo and evaluated on \emph{Waymo Eval}. }
    \label{fig:atleastone}%
\end{figure*}

Figure~\ref{fig:atleastone} is the twin of figure~\ref{fig:pointreidapproaches} from the main manuscript. Each figure plots the performance of our ReID networks on nuScenes (left) and Waymo (right) at different point densities. The difference between the two figures is seen on the x-axis. Figure~\ref{fig:atleastone} plots accuracy on pairs of observations with at least one observation containing $x$ points or more, while figure~\ref{fig:pointreidapproaches} requires that both observations in the pair have at least $x$ points. We observe that the increase in accuracy is steeper when restricting to cases where both observations have more than $x$ points. However, figure~\ref{fig:atleastone} shows the same trend but with a smaller slope and more samples for every $x$ value. 

\begin{figure*}[ht]
    \centering
    \subfloat{{\includegraphics[width=0.49\linewidth]{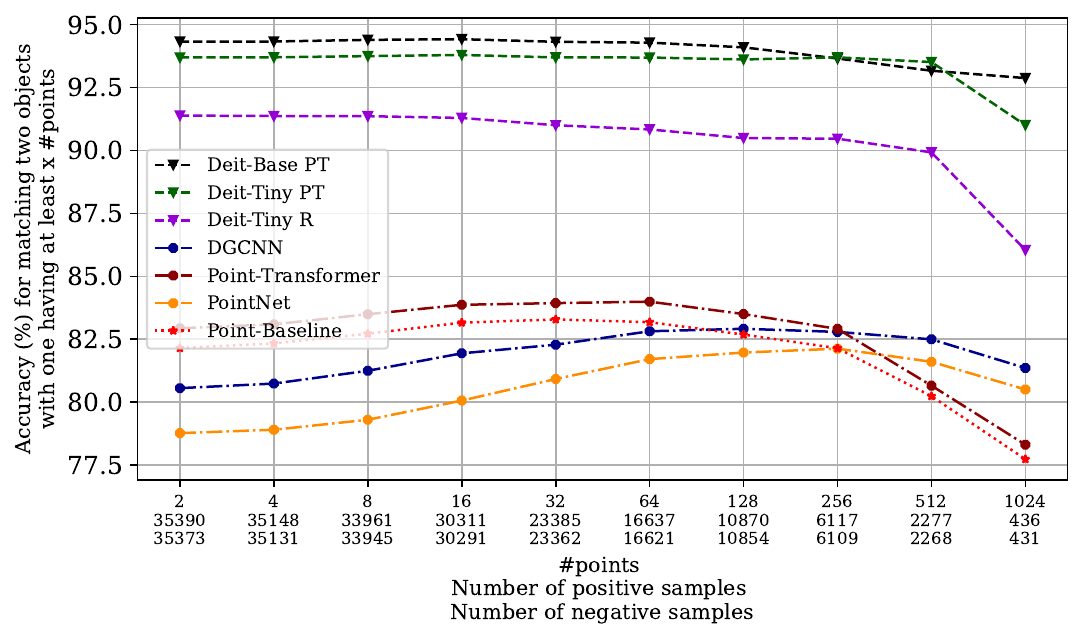} }}
    \subfloat{{\includegraphics[width=0.49\linewidth]{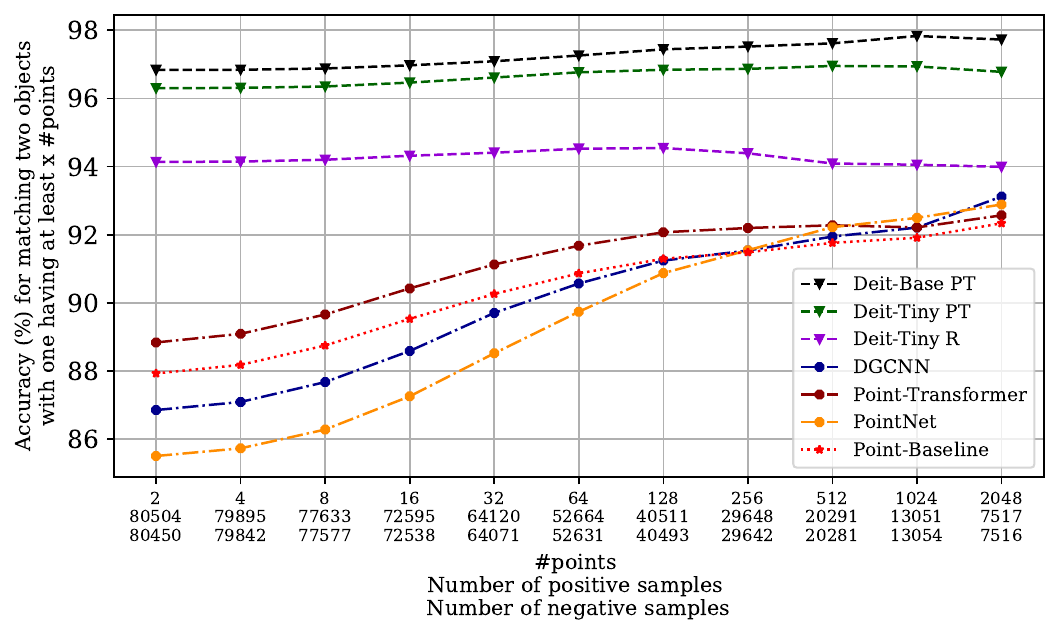} }}
    \vspace{-5pt}
    \caption{\textbf{Deformable v.s. Non-deformable objects.} The plot shows the performance of image and point cloud ReID networks for classes car and pedestrian as pairs of observations without at least one observation containing $x$ points are filtered out. We observe a similar trend for deformable and non-deformable objects: both increase with more points. However, we note that the increase in ReID performance for cars is steeper than for pedestrians. Left is evaluated on pedestrians from the \emph{Waymo Eval} set, while right is evaluated on cars from \emph{Waymo Eval}. }
    \vspace{-13pt}
    \label{fig:deformableone}
\end{figure*}

Similarly, figure~\ref{fig:deformableone} is the twin of figure \ref{fig:deformable} in the main manuscript. We see a similar trend of increased performance relative to the number of points but with a smaller slope compared to the plot which filters for both observations having more than $x$ points. Notably, performance improves faster on cars in figure \ref{fig:deformableone} than it does for pedestrians, while the opposite was true in figure \ref{fig:deformable}, suggesting that pairs of dense observations are needed for strong performance on deformable objects, while rigid objects can perform well with only one dense observation.

\section{Dataset visualization}
\label{sec:vis}
\begin{figure*}[ht]
    \centering
    \subfloat{{\includegraphics[width=0.9\linewidth]{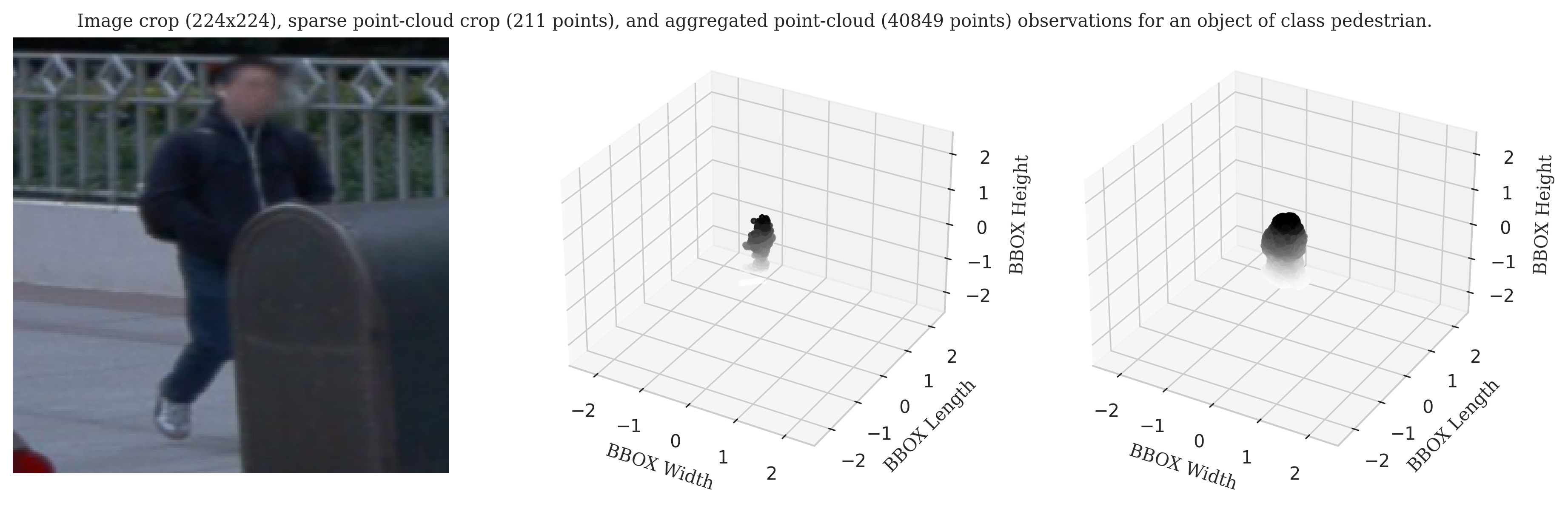} }}\\
    \subfloat{{\includegraphics[width=0.9\linewidth]{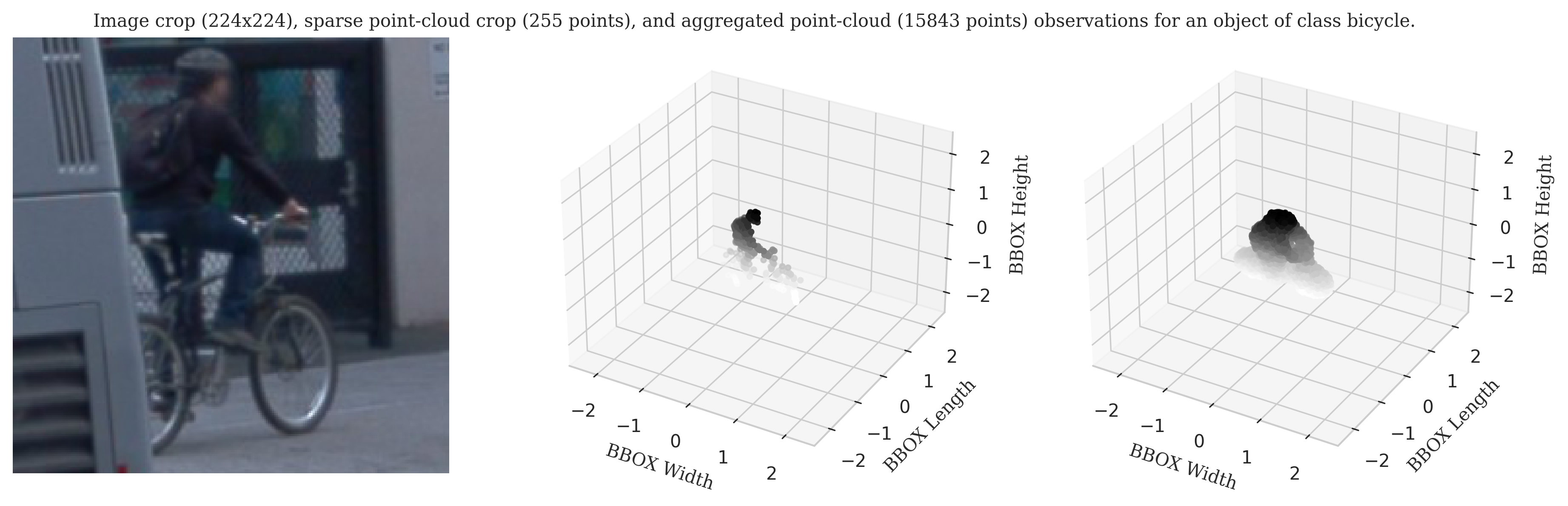} }}\\
    \subfloat{{\includegraphics[width=0.9\linewidth]{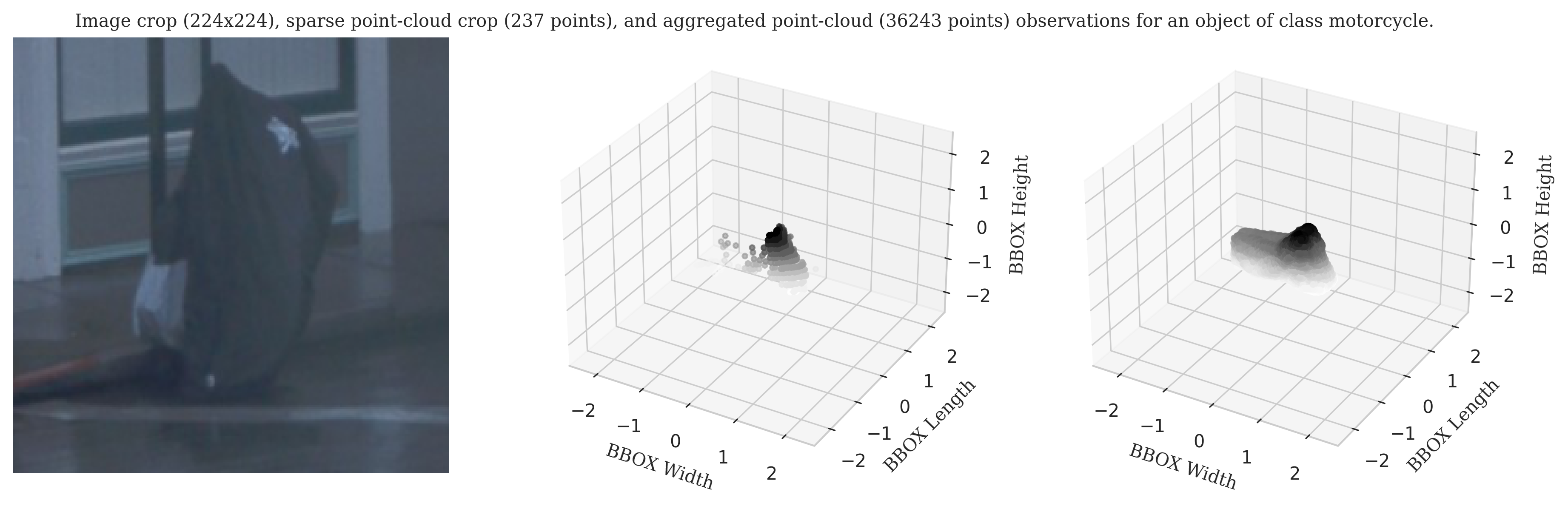} }}\\
    \caption{ \textbf{Samples from our Waymo ReID dataset for deformable objects.} The plot shows the cropped image (left) and cropped sparse point cloud (center) for the same predicted 3D bounding box (output by our centerpoint model). We also include the corresponding complete version of the point cloud (right), created by aggregating LiDAR scans over different observations and mirroring them about the object's center.}
    \label{fig:waymo-samples1}%
\end{figure*}

\begin{figure*}[ht]
    \centering
    \subfloat{{\includegraphics[width=0.9\linewidth]{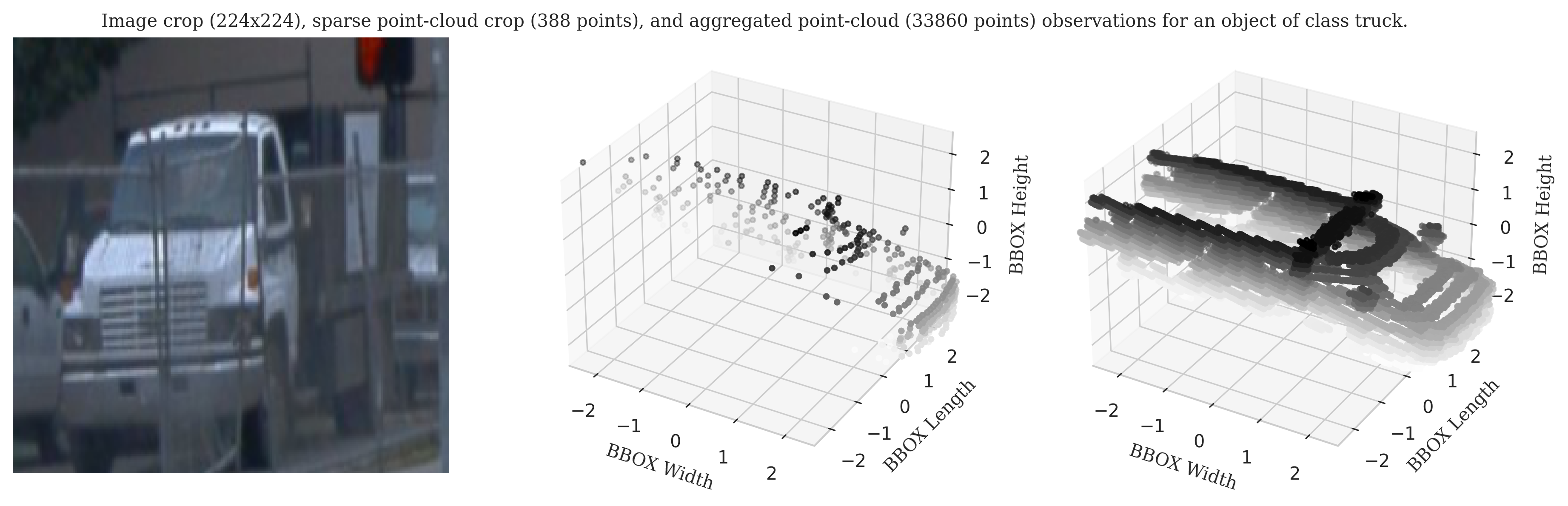} }}\\
    \subfloat{{\includegraphics[width=0.9\linewidth]{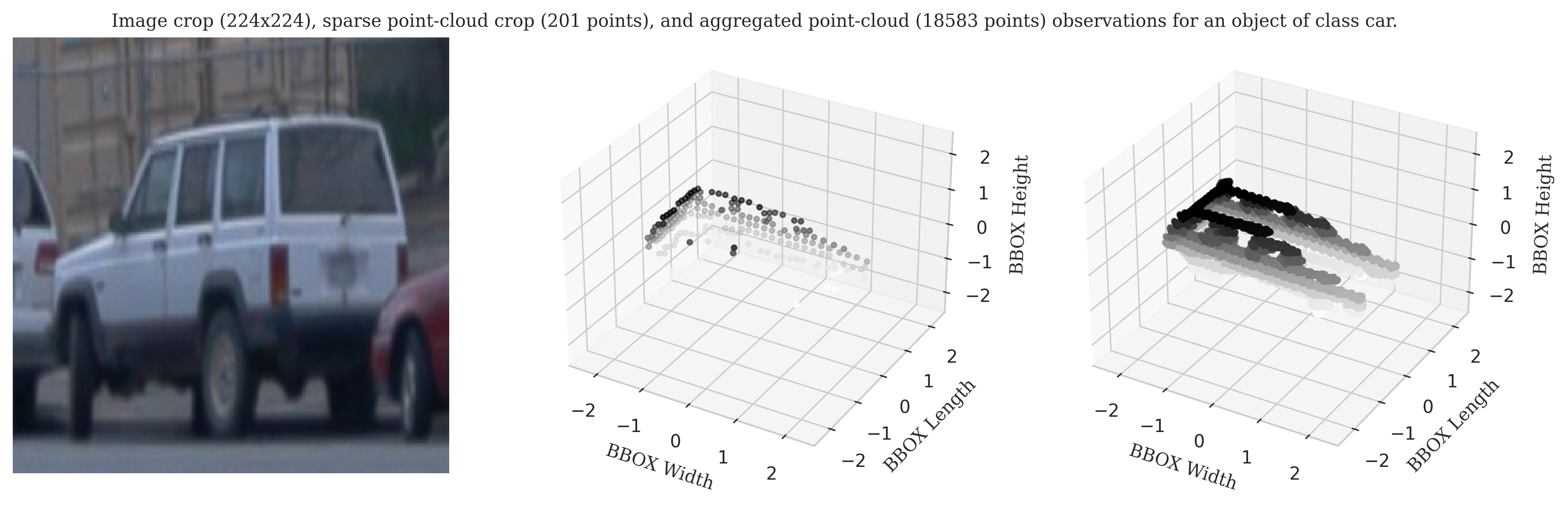} }}\\
    \subfloat{{\includegraphics[width=0.9\linewidth]{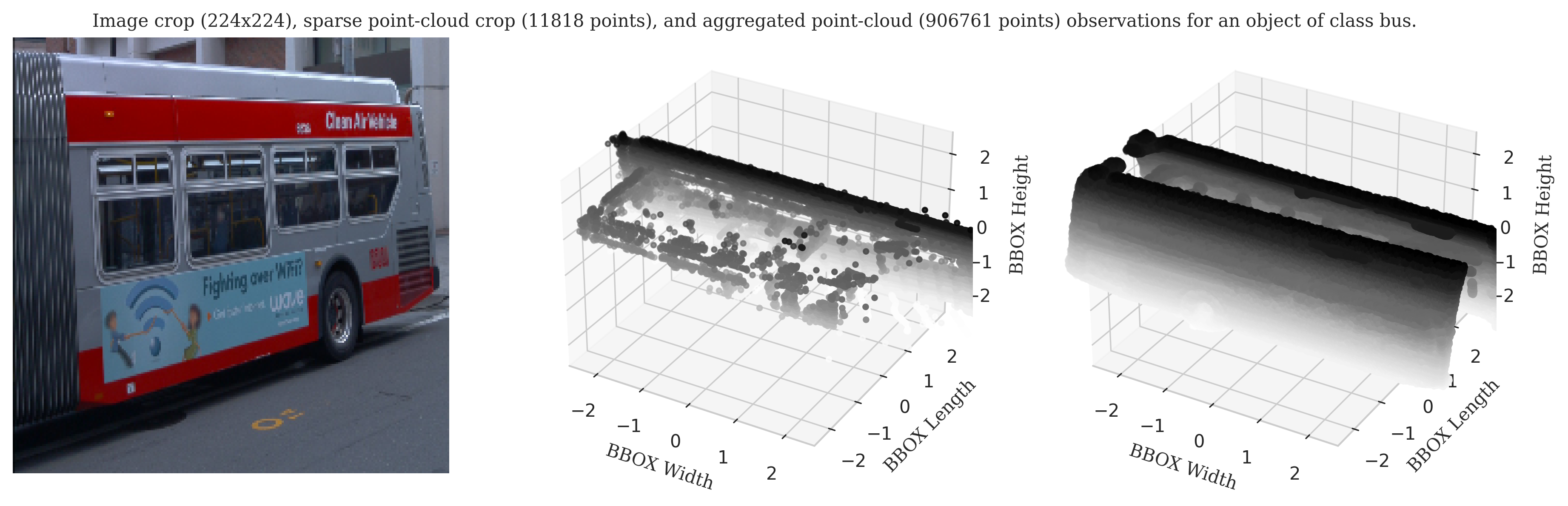} }}\\
    \caption{\textbf{Samples from our Waymo ReID dataset continued for rigid objects.} The plot shows the cropped image (left) and cropped sparse point cloud (center) for the same predicted 3D bounding box (output by our centerpoint model). We also include the corresponding complete version of the point cloud (right), created by aggregating LiDAR scans over different observations and mirroring them about the object's center.}
    \label{fig:waymo-samples2}%
\end{figure*}

In figures~\ref{fig:waymo-samples1} and \ref{fig:waymo-samples2}, we provide visual examples of samples from our Waymo ReID dataset. The images are cropped from projected 3D bounding boxes predicted by our centerpoint model (see sec. \ref{appendix-sec:detectors} for details). The center plot shows the LiDAR point associated to the same observation. We selected observations with more than $200$ points, so that it is possible for a human to make out the underlying object. However, most of the observations contain fewer than $200$ points. The leftmost plot shows complete point clouds created by aggregating the points from all observations using ground truth bounding boxes. Aggregated deformable objects (pedestrian, bicycle, and motorcycle) have a blob-like appearance, while rigid objects retain a more detailed shape. This is due to their deformability, causing them to take on many different poses over a sequence. 

\section{CFA block structure}
\label{sec-appendix:cfablock}
Our RTMM, illustrated in Fig.~\ref{fig:model}, compares two point clouds by symmetrically applying CFA blocks \cite{hui2022siamesepointtrans} between them. These are linear attention blocks, but with a few modifications; here we detail the exact structure used. CFA blocks receive two sets of points $\{\vx^{(i)}_1\}_i^{n_1} , \{\vx^{(i)}_2\}_i^{n_2}$, which we designate in stacked matrix form $\mX_1, \mX_2 \in \R^{n \times 3}$ henceforth, where the points are subsampled or resampled to size $n$ (we use $n=128$ for all our models), and their corresponding representations $f_\theta(\mX_1),f_\theta(\mX_2)$, where $f_\theta$ can be any set or sequence processing network. The block first computes linear cross-attention (LCA):
    \begin{align}
        \mL =& \text{ LCA}(f_\theta(\mX_1),f_\theta(\mX_2)+\mP,f_\theta(\mX_2)+\mP) 
    \end{align}
    where $\mP = \text{ MLP}_{pos}(\mX_2)$ is a positional encoding computed from the point cloud. Next, a layer normalization (LN) is applied followed by an MLP applied to the channel-wise concatenation of $\text{ LN}(\mL)$ and $f_\theta(\mX_1))$ and another layer normalization,
    \begin{align}
        \mL' =&  \text{ LN}(\text{ MLP}(\text{ LN}(\mL) \oplus f_\theta(\mX_1))).
    \end{align}
    Finally, a residual connection is applied to complete the CFA block:
    \begin{align}
         \text{CFA}(f_\theta(\mX_1),\mX_1,f_\theta(\mX_2),\mX_2) =& \mL' + f_\theta(\mX_1).
    \end{align}

\end{appendix}

\end{document}

%% file: math-commands.tex
\def\eqref#1{equation~\ref{#1}}

\def\1{\bm{1}}

\def\vx{{\bm{x}}}
\def\vy{{\bm{y}}}

\def\mA{{\bm{A}}}

\def\mL{{\bm{L}}}

\def\mN{{\bm{N}}}

\def\mP{{\bm{P}}}

\def\mX{{\bm{X}}}

\DeclareMathAlphabet{\mathsfit}{\encodingdefault}{\sfdefault}{m}{sl}
\SetMathAlphabet{\mathsfit}{bold}{\encodingdefault}{\sfdefault}{bx}{n}

\newcommand{\R}{\mathbb{R}}



%% file: main.bbl
\begin{thebibliography}{10}\itemsep=-1pt

\bibitem{alabdulmohsin2022scaling}
Ibrahim~M. Alabdulmohsin, Behnam Neyshabur, and Xiaohua Zhai.
\newblock Revisiting neural scaling laws in language and vision.
\newblock In {\em NeurIPS}, 2022.

\bibitem{ceasar2020nuscenes}
Holger Caesar, Varun Bankiti, Alex~H. Lang, Sourabh Vora, Venice~Erin Liong,
  Qiang Xu, Anush Krishnan, Yu Pan, Giancarlo Baldan, and Oscar Beijbom.
\newblock nuscenes: {A} multimodal dataset for autonomous driving.
\newblock In {\em 2020 {IEEE/CVF} Conference on Computer Vision and Pattern
  Recognition, {CVPR} 2020, Seattle, WA, USA, June 13-19, 2020}, pages
  11618--11628. Computer Vision Foundation / {IEEE}, 2020.

\bibitem{chang2019argoverse}
Ming{-}Fang Chang, John Lambert, Patsorn Sangkloy, Jagjeet Singh, Slawomir Bak,
  Andrew Hartnett, De Wang, Peter Carr, Simon Lucey, Deva Ramanan, and James
  Hays.
\newblock Argoverse: 3d tracking and forecasting with rich maps.
\newblock In {\em {IEEE} Conference on Computer Vision and Pattern Recognition,
  {CVPR} 2019, Long Beach, CA, USA, June 16-20, 2019}, pages 8748--8757.
  Computer Vision Foundation / {IEEE}, 2019.

\bibitem{choe2022pointmixer}
Jaesung Choe, Chunghyun Park, Fran{\c{c}}ois Rameau, Jaesik Park, and In~So
  Kweon.
\newblock Pointmixer: Mlp-mixer for point cloud understanding.
\newblock In Shai Avidan, Gabriel~J. Brostow, Moustapha Ciss{\'{e}},
  Giovanni~Maria Farinella, and Tal Hassner, editors, {\em Computer Vision -
  {ECCV} 2022 - 17th European Conference, Tel Aviv, Israel, October 23-27,
  2022, Proceedings, Part {XXVII}}, volume 13687 of {\em Lecture Notes in
  Computer Science}, pages 620--640. Springer, 2022.

\bibitem{chung2017stream}
Dahjung Chung, Khalid Tahboub, and Edward~J. Delp.
\newblock A two stream siamese convolutional neural network for person
  re-identification.
\newblock In {\em {IEEE} International Conference on Computer Vision, {ICCV}
  2017, Venice, Italy, October 22-29, 2017}, pages 1992--2000. {IEEE} Computer
  Society, 2017.

\bibitem{mmcv}
MMCV Contributors.
\newblock {MMCV: OpenMMLab} computer vision foundation.
\newblock \url{https://github.com/open-mmlab/mmcv}, 2018.

\bibitem{mmdet3d2020}
MMDetection3D Contributors.
\newblock {MMDetection3D: OpenMMLab} next-generation platform for general {3D}
  object detection.
\newblock \url{https://github.com/open-mmlab/mmdetection3d}, 2020.

\bibitem{deng2021vectorneuron}
Congyue Deng, Or Litany, Yueqi Duan, Adrien Poulenard, Andrea Tagliasacchi, and
  Leonidas~J. Guibas.
\newblock Vector neurons: {A} general framework for so(3)-equivariant networks.
\newblock In {\em 2021 {IEEE/CVF} International Conference on Computer Vision,
  {ICCV} 2021, Montreal, QC, Canada, October 10-17, 2021}, pages 12180--12189.
  {IEEE}, 2021.

\bibitem{eckstein2020large}
Viktor Eckstein, Arne Schumann, and Andreas Specker.
\newblock Large scale vehicle re-identification by knowledge transfer from
  simulated data and temporal attention.
\newblock In {\em 2020 {IEEE/CVF} Conference on Computer Vision and Pattern
  Recognition, {CVPR} Workshops 2020, Seattle, WA, USA, June 14-19, 2020},
  pages 2626--2631. Computer Vision Foundation / {IEEE}, 2020.

\bibitem{fuchs2020se3}
Fabian Fuchs, Daniel~E. Worrall, Volker Fischer, and Max Welling.
\newblock Se(3)-transformers: 3d roto-translation equivariant attention
  networks.
\newblock In Hugo Larochelle, Marc'Aurelio Ranzato, Raia Hadsell,
  Maria{-}Florina Balcan, and Hsuan{-}Tien Lin, editors, {\em Advances in
  Neural Information Processing Systems 33: Annual Conference on Neural
  Information Processing Systems 2020, NeurIPS 2020, December 6-12, 2020,
  virtual}, 2020.

\bibitem{gao2020complementary}
Cunyuan Gao, Yi Hu, Yi Zhang, Rui Yao, Yong Zhou, and Jiaqi Zhao.
\newblock Vehicle re-identification based on complementary features.
\newblock In {\em 2020 {IEEE/CVF} Conference on Computer Vision and Pattern
  Recognition, {CVPR} Workshops 2020, Seattle, WA, USA, June 14-19, 2020},
  pages 2520--2526. Computer Vision Foundation / {IEEE}, 2020.

\bibitem{ge2020afdet}
Runzhou Ge, Zhuangzhuang Ding, Yihan Hu, Yu Wang, Sijia Chen, Li Huang, and
  Yuan Li.
\newblock Afdet: Anchor free one stage 3d object detection.
\newblock {\em CoRR}, abs/2006.12671, 2020.

\bibitem{giancola2019completion}
Silvio Giancola, Jesus Zarzar, and Bernard Ghanem.
\newblock Leveraging shape completion for 3d siamese tracking.
\newblock In {\em {IEEE} Conference on Computer Vision and Pattern Recognition,
  {CVPR} 2019, Long Beach, CA, USA, June 16-20, 2019}, pages 1359--1368.
  Computer Vision Foundation / {IEEE}, 2019.

\bibitem{he2020multidomain}
Shuting He, Hao Luo, Weihua Chen, Miao Zhang, Yuqi Zhang, Fan Wang, Hao Li, and
  Wei Jiang.
\newblock Multi-domain learning and identity mining for vehicle
  re-identification.
\newblock In {\em 2020 {IEEE/CVF} Conference on Computer Vision and Pattern
  Recognition, {CVPR} Workshops 2020, Seattle, WA, USA, June 14-19, 2020},
  pages 2485--2493. Computer Vision Foundation / {IEEE}, 2020.

\bibitem{he2021transreid}
Shuting He, Hao Luo, Pichao Wang, Fan Wang, Hao Li, and Wei Jiang.
\newblock Transreid: Transformer-based object re-identification.
\newblock In {\em 2021 {IEEE/CVF} International Conference on Computer Vision,
  {ICCV} 2021, Montreal, QC, Canada, October 10-17, 2021}, pages 14993--15002.
  {IEEE}, 2021.

\bibitem{hermans2017defense}
Alexander Hermans, Lucas Beyer, and Bastian Leibe.
\newblock In defense of the triplet loss for person re-identification.
\newblock {\em CoRR}, abs/1703.07737, 2017.

\bibitem{hu2021hand}
Zheng Hu, Chuang Zhu, and Gang He.
\newblock Hard-sample guided hybrid contrast learning for unsupervised person
  re-identification.
\newblock In {\em 7th {IEEE} International Conference on Network Intelligence
  and Digital Content, {IC-NIDC} 2021, Beijing, China, November 17-19, 2021},
  pages 91--95. {IEEE}, 2021.

\bibitem{hui2021v2b}
Le Hui, Lingpeng Wang, Mingmei Cheng, Jin Xie, and Jian Yang.
\newblock 3d siamese voxel-to-bev tracker for sparse point clouds.
\newblock In Marc'Aurelio Ranzato, Alina Beygelzimer, Yann~N. Dauphin, Percy
  Liang, and Jennifer~Wortman Vaughan, editors, {\em Advances in Neural
  Information Processing Systems 34: Annual Conference on Neural Information
  Processing Systems 2021, NeurIPS 2021, December 6-14, 2021, virtual}, pages
  28714--28727, 2021.

\bibitem{hui2022siamesepointtrans}
Le Hui, Lingpeng Wang, Linghua Tang, Kaihao Lan, Jin Xie, and Jian Yang.
\newblock 3d siamese transformer network for single object tracking on point
  clouds.
\newblock In Shai Avidan, Gabriel~J. Brostow, Moustapha Ciss{\'{e}},
  Giovanni~Maria Farinella, and Tal Hassner, editors, {\em Computer Vision -
  {ECCV} 2022 - 17th European Conference, Tel Aviv, Israel, October 23-27,
  2022, Proceedings, Part {II}}, volume 13662 of {\em Lecture Notes in Computer
  Science}, pages 293--310. Springer, 2022.

\bibitem{jiao2022dynamic}
Bingliang Jiao, Lingqiao Liu, Liying Gao, Guosheng Lin, Lu Yang, Shizhou Zhang,
  Peng Wang, and Yanning Zhang.
\newblock Dynamically transformed instance normalization network for
  generalizable person re-identification.
\newblock In Shai Avidan, Gabriel~J. Brostow, Moustapha Ciss{\'{e}},
  Giovanni~Maria Farinella, and Tal Hassner, editors, {\em Computer Vision -
  {ECCV} 2022 - 17th European Conference, Tel Aviv, Israel, October 23-27,
  2022, Proceedings, Part {XIV}}, volume 13674 of {\em Lecture Notes in
  Computer Science}, pages 285--301. Springer, 2022.

\bibitem{jincloth2022}
Xin Jin, Tianyu He, Kecheng Zheng, Zhiheng Yin, Xu Shen, Zhen Huang, Ruoyu
  Feng, Jianqiang Huang, Zhibo Chen, and Xian{-}Sheng Hua.
\newblock Cloth-changing person re-identification from {A} single image with
  gait prediction and regularization.
\newblock In {\em {IEEE/CVF} Conference on Computer Vision and Pattern
  Recognition, {CVPR} 2022, New Orleans, LA, USA, June 18-24, 2022}, pages
  14258--14267. {IEEE}, 2022.

\bibitem{Katharopoulos2020linearattn}
Angelos Katharopoulos, Apoorv Vyas, Nikolaos Pappas, and Fran{\c{c}}ois
  Fleuret.
\newblock Transformers are rnns: Fast autoregressive transformers with linear
  attention.
\newblock In {\em Proceedings of the 37th International Conference on Machine
  Learning, {ICML} 2020, 13-18 July 2020, Virtual Event}, volume 119 of {\em
  Proceedings of Machine Learning Research}, pages 5156--5165. {PMLR}, 2020.

\bibitem{khorramshahi2020devil}
Pirazh Khorramshahi, Neehar Peri, Jun{-}Cheng Chen, and Rama Chellappa.
\newblock The devil is in the details: Self-supervised attention for vehicle
  re-identification.
\newblock In Andrea Vedaldi, Horst Bischof, Thomas Brox, and Jan{-}Michael
  Frahm, editors, {\em Computer Vision - {ECCV} 2020 - 16th European
  Conference, Glasgow, UK, August 23-28, 2020, Proceedings, Part {XIV}}, volume
  12359 of {\em Lecture Notes in Computer Science}, pages 369--386. Springer,
  2020.

\bibitem{kim2021eagermot}
Aleksandr Kim, Aljosa Osep, and Laura Leal{-}Taix{\'{e}}.
\newblock Eagermot: 3d multi-object tracking via sensor fusion.
\newblock In {\em {IEEE} International Conference on Robotics and Automation,
  {ICRA} 2021, Xi'an, China, May 30 - June 5, 2021}, pages 11315--11321.
  {IEEE}, 2021.

\bibitem{le2020lean}
Eric{-}Tuan Le, Iasonas Kokkinos, and Niloy~J. Mitra.
\newblock Going deeper with lean point networks.
\newblock In {\em 2020 {IEEE/CVF} Conference on Computer Vision and Pattern
  Recognition, {CVPR} 2020, Seattle, WA, USA, June 13-19, 2020}, pages
  9500--9509. Computer Vision Foundation / {IEEE}, 2020.

\bibitem{liciotti2016topview}
Daniele Liciotti, Marina Paolanti, Emanuele Frontoni, Adriano Mancini, and
  Primo Zingaretti.
\newblock Person re-identification dataset with {RGB-D} camera in a top-view
  configuration.
\newblock In Kamal Nasrollahi, Cosimo Distante, Gang Hua, Andrea Cavallaro,
  Thomas~B. Moeslund, Sebastiano Battiato, and Qiang Ji, editors, {\em Video
  Analytics. Face and Facial Expression Recognition and Audience Measurement -
  Third International Workshop, {VAAM} 2016, and Second International Workshop,
  {FFER} 2016, Cancun, Mexico, December 4, 2016, Revised Selected Papers},
  volume 10165 of {\em Lecture Notes in Computer Science}, pages 1--11.
  Springer, 2016.

\bibitem{zheng2019indentity}
Yutian Lin, Liang Zheng, Zhedong Zheng, Yu Wu, Zhilan Hu, Chenggang Yan, and Yi
  Yang.
\newblock Improving person re-identification by attribute and identity
  learning.
\newblock {\em Pattern Recognit.}, 95:151--161, 2019.

\bibitem{liu2017online}
Hong Liu, Liang Hu, and Liqian Ma.
\newblock Online {RGB-D} person re-identification based on metric model update.
\newblock {\em {CAAI} Trans. Intell. Technol.}, 2(1):48--55, 2017.

\bibitem{liu2020nonlocal}
Kai Liu, Zheng Xu, Zhaohui Hou, Zhicheng Zhao, and Fei Su.
\newblock Further non-local and channel attention networks for vehicle
  re-identification.
\newblock In {\em 2020 {IEEE/CVF} Conference on Computer Vision and Pattern
  Recognition, {CVPR} Workshops 2020, Seattle, WA, USA, June 14-19, 2020},
  pages 2494--2500. Computer Vision Foundation / {IEEE}, 2020.

\bibitem{liu2022bevfusion}
Zhijian Liu, Haotian Tang, Alexander Amini, Xinyu Yang, Huizi Mao, Daniela Rus,
  and Song Han.
\newblock Bevfusion: Multi-task multi-sensor fusion with unified bird's-eye
  view representation.
\newblock {\em CoRR}, abs/2205.13542, 2022.

\bibitem{loshchilov2019adamw}
Ilya Loshchilov and Frank Hutter.
\newblock Decoupled weight decay regularization.
\newblock In {\em 7th International Conference on Learning Representations,
  {ICLR} 2019, New Orleans, LA, USA, May 6-9, 2019}. OpenReview.net, 2019.

\bibitem{luo2019tricks}
Hao Luo, Youzhi Gu, Xingyu Liao, Shenqi Lai, and Wei Jiang.
\newblock Bag of tricks and a strong baseline for deep person
  re-identification.
\newblock In {\em {IEEE} Conference on Computer Vision and Pattern Recognition
  Workshops, {CVPR} Workshops 2019, Long Beach, CA, USA, June 16-20, 2019},
  pages 1487--1495. Computer Vision Foundation / {IEEE}, 2019.

\bibitem{ma2022residualmlp}
Xu Ma, Can Qin, Haoxuan You, Haoxi Ran, and Yun Fu.
\newblock Rethinking network design and local geometry in point cloud: {A}
  simple residual {MLP} framework.
\newblock In {\em The Tenth International Conference on Learning
  Representations, {ICLR} 2022, Virtual Event, April 25-29, 2022}.
  OpenReview.net, 2022.

\bibitem{munaro2014oneshot}
Matteo Munaro, Andrea Fossati, Alberto Basso, Emanuele Menegatti, and Luc~Van
  Gool.
\newblock One-shot person re-identification with a consumer depth camera.
\newblock In Shaogang Gong, Marco Cristani, Shuicheng Yan, and Chen~Change Loy,
  editors, {\em Person Re-Identification}, Advances in Computer Vision and
  Pattern Recognition, pages 161--181. Springer, 2014.

\bibitem{nagrani2021attentionbottleneck}
Arsha Nagrani, Shan Yang, Anurag Arnab, Aren Jansen, Cordelia Schmid, and Chen
  Sun.
\newblock Attention bottlenecks for multimodal fusion.
\newblock In Marc'Aurelio Ranzato, Alina Beygelzimer, Yann~N. Dauphin, Percy
  Liang, and Jennifer~Wortman Vaughan, editors, {\em Advances in Neural
  Information Processing Systems 34: Annual Conference on Neural Information
  Processing Systems 2021, NeurIPS 2021, December 6-14, 2021, virtual}, pages
  14200--14213, 2021.

\bibitem{patruno2019skeleton}
Cosimo Patruno, Roberto Marani, Grazia Cicirelli, Ettore Stella, and Tiziana
  D'Orazio.
\newblock People re-identification using skeleton standard posture and color
  descriptors from {RGB-D} data.
\newblock {\em Pattern Recognit.}, 89:77--90, 2019.

\bibitem{qipointnet2017}
Charles~Ruizhongtai Qi, Hao Su, Kaichun Mo, and Leonidas~J. Guibas.
\newblock Pointnet: Deep learning on point sets for 3d classification and
  segmentation.
\newblock In {\em 2017 {IEEE} Conference on Computer Vision and Pattern
  Recognition, {CVPR} 2017, Honolulu, HI, USA, July 21-26, 2017}, pages 77--85.
  {IEEE} Computer Society, 2017.

\bibitem{qi2017pnpp}
Charles~Ruizhongtai Qi, Li Yi, Hao Su, and Leonidas~J. Guibas.
\newblock Pointnet++: Deep hierarchical feature learning on point sets in a
  metric space.
\newblock In Isabelle Guyon, Ulrike von Luxburg, Samy Bengio, Hanna~M. Wallach,
  Rob Fergus, S.~V.~N. Vishwanathan, and Roman Garnett, editors, {\em Advances
  in Neural Information Processing Systems 30: Annual Conference on Neural
  Information Processing Systems 2017, December 4-9, 2017, Long Beach, CA,
  {USA}}, pages 5099--5108, 2017.

\bibitem{qi2020p2b}
Haozhe Qi, Chen Feng, Zhiguo Cao, Feng Zhao, and Yang Xiao.
\newblock {P2B:} point-to-box network for 3d object tracking in point clouds.
\newblock In {\em 2020 {IEEE/CVF} Conference on Computer Vision and Pattern
  Recognition, {CVPR} 2020, Seattle, WA, USA, June 13-19, 2020}, pages
  6328--6337. Computer Vision Foundation / {IEEE}, 2020.

\bibitem{qian2022decoupling}
Wen Qian, Hao Luo, Silong Peng, Fan Wang, Chen Chen, and Hao Li.
\newblock Unstructured feature decoupling for vehicle re-identification.
\newblock In Shai Avidan, Gabriel~J. Brostow, Moustapha Ciss{\'{e}},
  Giovanni~Maria Farinella, and Tal Hassner, editors, {\em Computer Vision -
  {ECCV} 2022 - 17th European Conference, Tel Aviv, Israel, October 23-27,
  2022, Proceedings, Part {XIV}}, volume 13674 of {\em Lecture Notes in
  Computer Science}, pages 336--353. Springer, 2022.

\bibitem{sebastian2020dual}
Clint Sebastian, Raffaele Imbriaco, Egor Bondarev, and Peter H.~N. de With.
\newblock Dual embedding expansion for vehicle re-identification.
\newblock In {\em 2020 {IEEE/CVF} Conference on Computer Vision and Pattern
  Recognition, {CVPR} Workshops 2020, Seattle, WA, USA, June 14-19, 2020},
  pages 2475--2484. Computer Vision Foundation / {IEEE}, 2020.

\bibitem{shan2022sotwt}
Jiayao Shan, Sifan Zhou, Yubo Cui, and Zheng Fang.
\newblock Real-time 3d single object tracking with transformer.
\newblock {\em CoRR}, abs/2209.00860, 2022.

\bibitem{smith2019super}
Leslie~N Smith and Nicholay Topin.
\newblock Super-convergence: Very fast training of neural networks using large
  learning rates.
\newblock In {\em Artificial intelligence and machine learning for multi-domain
  operations applications}, volume 11006, pages 369--386. SPIE, 2019.

\bibitem{alphaprimelidar}
Autonomous Stuff.
\newblock Alpha prime, powering safe autonomy.

\bibitem{sun2020waymo}
Pei Sun, Henrik Kretzschmar, Xerxes Dotiwalla, Aurelien Chouard, Vijaysai
  Patnaik, Paul Tsui, James Guo, Yin Zhou, Yuning Chai, Benjamin Caine, Vijay
  Vasudevan, Wei Han, Jiquan Ngiam, Hang Zhao, Aleksei Timofeev, Scott
  Ettinger, Maxim Krivokon, Amy Gao, Aditya Joshi, Yu Zhang, Jonathon Shlens,
  Zhifeng Chen, and Dragomir Anguelov.
\newblock Scalability in perception for autonomous driving: Waymo open dataset.
\newblock In {\em 2020 {IEEE/CVF} Conference on Computer Vision and Pattern
  Recognition, {CVPR} 2020, Seattle, WA, USA, June 13-19, 2020}, pages
  2443--2451. Computer Vision Foundation / {IEEE}, 2020.

\bibitem{touvron2021deit}
Hugo Touvron, Matthieu Cord, Matthijs Douze, Francisco Massa, Alexandre
  Sablayrolles, and Herv{\'{e}} J{\'{e}}gou.
\newblock Training data-efficient image transformers {\&} distillation through
  attention.
\newblock In Marina Meila and Tong Zhang, editors, {\em Proceedings of the 38th
  International Conference on Machine Learning, {ICML} 2021, 18-24 July 2021,
  Virtual Event}, volume 139 of {\em Proceedings of Machine Learning Research},
  pages 10347--10357. {PMLR}, 2021.

\bibitem{wang2018granularity}
Guanshuo Wang, Yufeng Yuan, Xiong Chen, Jiwei Li, and Xi Zhou.
\newblock Learning discriminative features with multiple granularities for
  person re-identification.
\newblock In Susanne Boll, Kyoung~Mu Lee, Jiebo Luo, Wenwu Zhu, Hyeran Byun,
  Chang~Wen Chen, Rainer Lienhart, and Tao Mei, editors, {\em 2018 {ACM}
  Multimedia Conference on Multimedia Conference, {MM} 2018, Seoul, Republic of
  Korea, October 22-26, 2018}, pages 274--282. {ACM}, 2018.

\bibitem{wangcamomot2021}
Li Wang, Xinyu Zhang, Wenyuan Qin, Xiaoyu Li, Lei Yang, Zhiwei Li, Lei Zhu,
  Hong Wang, Jun Li, and Huaping Liu.
\newblock {CAMO-MOT:} combined appearance-motion optimization for 3d
  multi-object tracking with camera-lidar fusion.
\newblock {\em CoRR}, abs/2209.02540, 2022.

\bibitem{wang2020cen}
Yikai Wang, Wenbing Huang, Fuchun Sun, Tingyang Xu, Yu Rong, and Junzhou Huang.
\newblock Deep multimodal fusion by channel exchanging.
\newblock In Hugo Larochelle, Marc'Aurelio Ranzato, Raia Hadsell,
  Maria{-}Florina Balcan, and Hsuan{-}Tien Lin, editors, {\em Advances in
  Neural Information Processing Systems 33: Annual Conference on Neural
  Information Processing Systems 2020, NeurIPS 2020, December 6-12, 2020,
  virtual}, 2020.

\bibitem{wang2019dgcnn}
Yue Wang, Yongbin Sun, Ziwei Liu, Sanjay~E. Sarma, Michael~M. Bronstein, and
  Justin~M. Solomon.
\newblock Dynamic graph {CNN} for learning on point clouds.
\newblock {\em {ACM} Trans. Graph.}, 38(5):146:1--146:12, 2019.

\bibitem{wei2020retail}
Yuchen Wei, Son~N. Tran, Shuxiang Xu, Byeong~Ho Kang, and Matthew Springer.
\newblock Deep learning for retail product recognition: Challenges and
  techniques.
\newblock {\em Comput. Intell. Neurosci.}, 2020:8875910:1--8875910:23, 2020.

\bibitem{weng2020gnn3dmot}
Xinshuo Weng, Yongxin Wang, Yunze Man, and Kris~M. Kitani.
\newblock {GNN3DMOT:} graph neural network for 3d multi-object tracking with
  2d-3d multi-feature learning.
\newblock In {\em 2020 {IEEE/CVF} Conference on Computer Vision and Pattern
  Recognition, {CVPR} 2020, Seattle, WA, USA, June 13-19, 2020}, pages
  6498--6507. Computer Vision Foundation / {IEEE}, 2020.

\bibitem{willes2022intertrack}
John Willes, Cody Reading, and Steven~L. Waslander.
\newblock Intertrack: Interaction transformer for 3d multi-object tracking.
\newblock {\em CoRR}, abs/2208.08041, 2022.

\bibitem{yin2021centerpoint}
Tianwei Yin, Xingyi Zhou, and Philipp Kr{\"{a}}henb{\"{u}}hl.
\newblock Center-based 3d object detection and tracking.
\newblock In {\em {IEEE} Conference on Computer Vision and Pattern Recognition,
  {CVPR} 2021, virtual, June 19-25, 2021}, pages 11784--11793. Computer Vision
  Foundation / {IEEE}, 2021.

\bibitem{Zheng2018picknplace}
Andy Zeng, Shuran Song, Kuan{-}Ting Yu, Elliott Donlon, Francois~Robert Hogan,
  Maria Bauz{\'{a}}, Daolin Ma, Orion Taylor, Melody Liu, Eudald Romo, Nima
  Fazeli, Ferran Alet, Nikhil~Chavan Dafle, Rachel~M. Holladay, Isabella
  Morona, Prem~Qu Nair, Druck Green, Ian~H. Taylor, Weber Liu, Thomas~A.
  Funkhouser, and Alberto Rodriguez.
\newblock Robotic pick-and-place of novel objects in clutter with
  multi-affordance grasping and cross-domain image matching.
\newblock In {\em 2018 {IEEE} International Conference on Robotics and
  Automation, {ICRA} 2018, Brisbane, Australia, May 21-25, 2018}, pages 1--8.
  {IEEE}, 2018.

\bibitem{zeng2020hardbatch}
Kaiwei Zeng, Munan Ning, Yaohua Wang, and Yang Guo.
\newblock Hierarchical clustering with hard-batch triplet loss for person
  re-identification.
\newblock In {\em 2020 {IEEE/CVF} Conference on Computer Vision and Pattern
  Recognition, {CVPR} 2020, Seattle, WA, USA, June 13-19, 2020}, pages
  13654--13662. Computer Vision Foundation / {IEEE}, 2020.

\bibitem{zhao2021pointtransformer}
Hengshuang Zhao, Li Jiang, Jiaya Jia, Philip H.~S. Torr, and Vladlen Koltun.
\newblock Point transformer.
\newblock In {\em 2021 {IEEE/CVF} International Conference on Computer Vision,
  {ICCV} 2021, Montreal, QC, Canada, October 10-17, 2021}, pages 16239--16248.
  {IEEE}, 2021.

\bibitem{zhao2021hausdorff}
Jianan Zhao, Fengliang Qi, Guangyu Ren, and Lin Xu.
\newblock Phd learning: Learning with pompeiu-hausdorff distances for
  video-based vehicle re-identification.
\newblock In {\em {IEEE} Conference on Computer Vision and Pattern Recognition,
  {CVPR} 2021, virtual, June 19-25, 2021}, pages 2225--2235. Computer Vision
  Foundation / {IEEE}, 2021.

\bibitem{zheng2019consistent}
Meng Zheng, Srikrishna Karanam, Ziyan Wu, and Richard~J. Radke.
\newblock Re-identification with consistent attentive siamese networks.
\newblock In {\em {IEEE} Conference on Computer Vision and Pattern Recognition,
  {CVPR} 2019, Long Beach, CA, USA, June 16-20, 2019}, pages 5735--5744.
  Computer Vision Foundation / {IEEE}, 2019.

\bibitem{zheng2020beyond}
Zhedong Zheng, Minyue Jiang, Zhigang Wang, Jian Wang, Zechen Bai, Xuanmeng
  Zhang, Xin Yu, Xiao Tan, Yi Yang, Shilei Wen, and Errui Ding.
\newblock Going beyond real data: {A} robust visual representation for vehicle
  re-identification.
\newblock In {\em 2020 {IEEE/CVF} Conference on Computer Vision and Pattern
  Recognition, {CVPR} Workshops 2020, Seattle, WA, USA, June 14-19, 2020},
  pages 2550--2558. Computer Vision Foundation / {IEEE}, 2020.

\bibitem{zheng2021vehiclenet}
Zhedong Zheng, Tao Ruan, Yunchao Wei, Yi Yang, and Tao Mei.
\newblock Vehiclenet: Learning robust visual representation for vehicle
  re-identification.
\newblock {\em {IEEE} Trans. Multim.}, 23:2683--2693, 2021.

\bibitem{zheng20223dreid}
Zhedong Zheng, Xiaohan Wang, Nenggan Zheng, and Yi Yang.
\newblock Parameter-efficient person re-identification in the 3d space.
\newblock {\em IEEE Transactions on Neural Networks and Learning Systems},
  pages 1--14, 2022.

\bibitem{zheng2019joint}
Zhedong Zheng, Xiaodong Yang, Zhiding Yu, Liang Zheng, Yi Yang, and Jan Kautz.
\newblock Joint discriminative and generative learning for person
  re-identification.
\newblock In {\em {IEEE} Conference on Computer Vision and Pattern Recognition,
  {CVPR} 2019, Long Beach, CA, USA, June 16-20, 2019}, pages 2138--2147.
  Computer Vision Foundation / {IEEE}, 2019.

\bibitem{zheng2018cnn}
Zhedong Zheng, Liang Zheng, and Yi Yang.
\newblock A discriminatively learned {CNN} embedding for person
  reidentification.
\newblock {\em {ACM} Trans. Multim. Comput. Commun. Appl.}, 14(1):13:1--13:20,
  2018.

\bibitem{zhu2020vocreid}
Xiangyu Zhu, Zhenbo Luo, Pei Fu, and Xiang Ji.
\newblock Voc-reid: Vehicle re-identification based on
  vehicle-orientation-camera.
\newblock {\em CoRR}, abs/2004.09164, 2020.

\end{thebibliography}
